\documentclass{article}
\usepackage{dsfont}

\usepackage{amsmath}
\usepackage{graphicx}
\usepackage{amssymb}
\usepackage{amsthm}
\usepackage{caption}
\usepackage{subcaption}
\usepackage{bbm}
\usepackage{mathtools}
\usepackage{algorithmic}
\usepackage{courier} 
\usepackage{algorithm}
\graphicspath{{./figures/}}
\usepackage[round]{natbib}
\renewcommand{\bibname}{References}
\renewcommand{\bibsection}{\subsubsection*{\bibname}}

\usepackage{chngcntr}
\usepackage{apptools}
\AtAppendix{\counterwithin{thm}{section}}

\usepackage{hyperref}
\usepackage{xcolor}

\usepackage{mathtools}

\newcommand{\norm}[1]{\Vert #1 \Vert}
\numberwithin{equation}{section}
\newcommand{\ee}{{\rm e}\hspace{1pt}}

\newcommand{\dd}{\hspace{1pt}{\rm d}\hspace{0.5pt}}
\newcommand{\abs}[1]{\left| #1 \right|}
\newcommand{\veps}{\varepsilon}

\newcommand{\wh}{\widehat}
\newtheorem{thm}{Theorem}
\newtheorem{lem}[thm]{Lemma}
\newtheorem{cor}[thm]{Corollary}
\newtheorem{defn}[thm]{Definition}

\newcommand{\M}{\mathcal{M}}

\newcommand{\wt}[1]{\widetilde{#1}}

\title{Auditing Differential Privacy Guarantees \\ Using Density Estimation 
}

\author{Antti Koskela and Jafar Mohammadi \vspace{5mm} \\
Nokia Bell Labs \\
 }

\date{}

\usepackage{graphicx}

\begin{document}

\maketitle

\begin{abstract}We present a novel method for accurately auditing the differential privacy (DP) guarantees of DP mechanisms. In particular, our solution is applicable to auditing DP guarantees of machine learning (ML) models. Previous auditing methods tightly capture the privacy guarantees of DP-SGD trained models in the white-box setting where the auditor has access to all intermediate models; however, the success of these methods depends on a priori information about the parametric form of the noise and the subsampling ratio used for sampling the gradients. 
We present a method that does not require such information and is agnostic to the randomization used for the underlying mechanism. 
Similarly to several previous DP auditing methods, we assume that the auditor has access to a set of independent observations from two one-dimensional distributions corresponding to outputs from two neighbouring datasets. 
Furthermore, our solution is based on a simple histogram-based density estimation technique to find lower bounds for the statistical distance between these distributions when measured using the hockey-stick divergence.
We show that our approach also naturally generalizes the previously considered class of threshold membership inference auditing methods. We improve upon accurate auditing methods such as the $f$-DP auditing. Moreover, we address an open problem on how to accurately audit the subsampled Gaussian mechanism without any knowledge of the parameters of the underlying mechanism.

\end{abstract}

\section{Introduction} 

Differential Privacy (DP)~\cite{dwork_et_al_2006} limits the disclosure of membership information of individuals in statistical data analysis. It has been successfully applied also to the training of machine learning (ML) models, where the de facto standard is the DP stochastic gradient descent (DP-SGD)~\cite{song2013stochastic,Abadi2016}. DP-SGD enables the analysis of formal $(\veps, \delta)$-DP guarantees via \textit{composition analysis} in a threat model where the guarantees hold against an adversary that has the access to the whole history of models. 
Using modern numerical accounting tools~\cite{koskela2020,zhu2021optimal,gopi2021}, it is also possible to obtain accurate $(\veps, \delta)$-DP guarantees for DP-SGD in this threat model.

We motivate the privacy auditing problem by the following scenario. Consider a federated learning (FL) setup, where a non-fully-trusted server participates in enhancing the DP protection by aggregating the local model updates and adding noise to the global updates. In order to achieve the theoretical privacy guarantees of DP-SGD, an overall notoriously difficult implementation setup is needed~\cite{tramer2022debugging,nasr2023tight}. Since parts of the model updates are performed by an external entity, there is no full certainty for a data-owner that the DP guarantees hold~\cite{maddock2023canife,andrew2024one}. This raises the question: how could the data owner conduct privacy auditing to ensure at least a certain amount of DP protection? Therefore, establishing some certainty about lower bounds for the DP parameters $\veps$ and $\delta$ is essential.

The problem of DP auditing has increasingly gained attention during recent years. Many of the existing works on DP-SGD auditing focus on inserting well-designed data elements or gradients into the training dataset, coined as the \textit{canaries}. By observing their effect later in the trained model one can infer about the DP guarantees~\cite{jagielski2020auditing,nasr2021adversary,pillutla2023unleashing,nasr2023tight}. We note that these methods commonly also require training several models in order to obtain the estimates of the DP guarantees, even up to thousands~\cite{pillutla2023unleashing}. 
To overcome the computational burden of training the model multiple times, recently~\cite{steinke2023privacy} and~\cite{andrew2024one} have proposed different approaches where auditing can be carried out in a single DP-SGD training iteration.  

Most of the methods such as~\cite{jagielski2020auditing,nasr2021adversary,nasr2023tight,pillutla2023unleashing,andrew2024one} 
are ultimately based on estimating the $(\veps,\delta)$-DP distance between two distributions that correspond to the outcomes of DP mechanisms originating from datasets that differ only by one data element. For instance, in black-box auditing, one distribution would correspond to loss function values evaluated on a dataset including a given data sample $z$ and the other one on the same dataset with $z$ excluded~\cite{jagielski2020auditing,nasr2021adversary,nasr2023tight}.
The $(\veps,\delta)$-guarantees are then commonly estimated using threshold membership inference attacks~\cite{yeom2018privacy,carlini2022membership}, where a model is deemed to contain the given sample in case its loss function value for that sample is below a certain threshold value. 
By considering multiple models trained once with and once without one differing sample and by measuring the false positive rates (FPRs) and false negative rates (FNRs) of the membership inferences, empirical $\veps$-estimates can be derived for a given value of $\delta$~\cite{kairouz2015composition}. 
Our work can be seen as a generalization of this approach such that
we estimate the two neighboring distributions using histograms. As we show, that the empirical $\veps$-values given by the threshold membership inference attacks are equivalent to measuring the hockey-stick divergence between the two discrete estimates obtained by histogram estimation of the distributions with two bins determined by the threshold value. 

One drawback of the threshold membership inference based auditing methods is that they tend to overly underestimate the $\veps$-values. To this end,~\cite{nasr2023tight} proposes $f$-DP auditing, where a certain trade-off curve $f$ is fitted such that the FNRs and FPRs of the membership inference are inferred to stay below $f$ with high confidence, thus leading to high-confidence lower bounds for the DP parameters $\veps$ and $\delta$. This approach, however, also has its drawbacks, as its success depends on having suitable candidate $f$ for the trade-off curve which generally requires
a priori information about the DP randomization mechanism. In addition, it often involves a complicated numerical integration procedure, which may lead to instabilities that we also demonstrate in this paper. Our approach is similar in the sense that we also aim to accurately approximate the trade-off function. Our approach differs in that its success does not depend on any a priori information about the DP mechanism and it has a simple and robust implementation.

In the FL setting there are two notable works related to ours, both of which are based on estimating the $(\veps,\delta)$-distance between two Gaussian distributions~\cite{andrew2024one,maddock2023canife}. The work~\cite{andrew2024one} advocates inserting randomly sampled canaries in the model updates and the method given in~\cite{maddock2023canife} is based on carefully crafting canary gradients. 
We show that our approach can be used to generalize the auditing method of~\cite{andrew2024one} for cases where the auditor does not have a priori information available about the noise used for randomizing model updates. We analytically show that our histogram-based method gives asymptotically the correct guarantees as the model dimension increases, similarly to the method of~\cite{andrew2024one} which uses a priori information about the noise.


Our work touches upon several challenging research areas, including density estimation and confidence interval estimation for discrete distributions. For each area, we rely on relatively simple benchmark results, as the primary focus of our work is to introduce a new concept for auditing a DP mechanisms.

Our paper is organized as follows. After presenting the necessary definitions and results on DP, in Section~\ref{sec:scores}, we describe the idea of obtaining $(\veps,\delta)$-DP lower bounds via the hockey-stick divergence between certain histograms-estimates. Then, we give some numerical examples to illustrate the benefits of our approach in Section~\ref{sec:num_examples}.
In Section~\ref{sec:relation}, we sketch the way that our approach generalizes the threshold inference auditing, and 
in Section~\ref{sec:tv}, we analytically illustrate that the total variation distance leads to a robust estimator in case the privacy profiles depend on a single parameter. 
In Section~\ref{sec:oneshot}, we  show that the density estimation based approach can also generalize the one-shot auditing method of~\cite{andrew2024one} that uses random gradient canaries.
Lastly, experiments of Section~\ref{sec:experiments} on a small neural network in both black-box and white-box settings further confirm the benefits of the density estimation approach for DP auditing.

Our main contributions can be summarized as follows:
\begin{itemize}

    \item We introduce a novel method of auditing the DP guarantees using samples from distributions that are a priori known to be $(\veps,\delta)$-close. This scenario fits perfectly to several previously considered black-box and white-box auditing scenarios~\cite{jagielski2020auditing,nasr2021adversary,nasr2023tight}. Our method expands the class of threshold membership inference methods by simultaneously considering several membership regions and by using histogram estimation of the distributions to obtain more accurate estimate of the $(\veps,\delta)$-distance.

    \item We solve an open problem posed in~\cite{nasr2023tight} on how to tightly audit the subsampled Gaussian mechanism. As the example given in~\cite{nasr2023tight} shows, a single threshold membership inference is not able to capture the accurate trade-off curve of the subsampled Gaussian mechanism. We show, both theoretically and empirically, that the trade-off curve estimated using our solution converges to the accurate trade-off curve.
    
    \item We demonstrate that in case the trade-off curve is defined by a single parameter, combining the histogram-based approach with the total variation distance yields the most robust estimate of the trade-off curve.

    \item We propose a heuristic algorithm for estimating the privacy loss distribution of the underlying mechanism in the white-box auditing setting. This enables obtaining accurate estimates for a given number of compositions of the DP mechanisms to be audited.

    \item We perform numerical experiments on neural network training on a benchmark dataset to illustrate the benefits of our approach in both black-box and white-box auditing scenarios.

\end{itemize}

\section{Background} \label{sec:dp}

\subsection{Differential Privacy}

We denote the space of possible data points by $X$. We denote a dataset containing $n$ data points as $D = (x_1,\ldots,x_n) \in X^n$, and the space of all possible datasets
(of all sizes) by $\mathcal{X}$. We say $D$ and $D'$ are neighboring datasets if we get one by substituting one element in the other. 
We say that a mechanism $\mathcal{M} \, : \, \mathcal{X} \rightarrow \mathcal{O}$ is $(\veps,\delta)$-DP if the output distributions for neighboring datasets are always
$(\veps,\delta)$-indistinguishable.

\begin{defn} \label{def:dp}
	Let $\varepsilon \ge 0$ and $\delta \in [0,1]$.	Mechanism $\mathcal{M} \, : \, \mathcal{X} \rightarrow \mathcal{O}$ 
 is $(\veps, \delta)$-DP	if for every pair of neighboring datasets $D,D' \in \mathcal{X}$ and every measurable set $E \subset \mathcal{O}$, 
	$$
	\mathbb{P}( \mathcal{M}(D) \in E ) \leq \ee^\varepsilon \mathbb{P} (\mathcal{M}(D') \in E ) + \delta. 
	$$
\end{defn}
The tight $(\veps,\delta)$-guarantees for a mechanism $\M$ can be stated using the hockey-stick divergence. 
For $\alpha \geq 0$ the hockey-stick divergence $H_{\alpha}$ from a distribution $P$ to a distribution $Q$ is defined as
\begin{equation} \label{eq:alphadiv}
	H_\alpha(P||Q) = \int \left[P(t) - \alpha \cdot Q(t) \right]_+ \, \dd t,
\end{equation}
where $[t]_+ = \max\{0,t\}$. 
The $(\veps,\delta)$-DP guarantees can be characterized using the hockey-stick divergence as follows (see Theorem 1 in~\cite{balle2018subsampling}).
\begin{lem} \label{lem:zhu_lemma5}
For a given $\veps \in \mathbb{R}$, a mechanism $\mathcal{M}$ satisfies $(\epsilon,\delta)$-DP \emph{if and only if},  for all neighboring datasets $D,D'$,
$$
H_{\ee^\veps}(\mathcal{M}(D)||\mathcal{M}(D')) \leq \delta.
$$
\end{lem}
We also refer to $\delta_{\mathcal{M}}(\veps) := \max_{X \sim X'} H_{e^\epsilon}(\mathcal{M}(X)||\mathcal{M}(X'))$
as the \emph{privacy profile} of mechanism $\mathcal{M}$. 

By Lemma~\ref{lem:zhu_lemma5}, if we can bound $H_{\ee^\veps}(\mathcal{M}(D)||\mathcal{M}(D'))$ accurately for all neighboring datasets $D,D'$, we also obtain accurate  $(\veps,\delta)$-DP bounds. For compositions of general DP mechanisms, this can be carried out by using so-called dominating pairs of distributions~\cite{zhu2021optimal} and numerical techniques~\cite{koskela2021tight,gopi2021}.
In some cases, such as for the Gaussian mechanism, the hockey-stick divergence~\eqref{eq:alphadiv} leads to analytical expressions
for tight $(\veps,\delta)$-DP guarantees~\cite{balle2018gauss}.

\begin{lem} \label{lem:gauss_dp}
Let $d_0,d_1\in \mathbb{R}^d$, $\sigma \geq 0$, and let $P$ be the density function of $\mathcal{N}(d_0,\sigma^2 I_d)$ and $Q$ the density function of $\mathcal{N}(d_1,\sigma^2 I_d)$.
Then, for all  $\veps \in \mathbb{R}$, the divergence $H_{\ee^\veps}(P || Q)$ is given by the expression
\begin{equation} \label{eq:delta_gaussian}
    \delta(\veps) = \Phi\left( - \frac{\veps\sigma}{\Delta} + \frac{\Delta}{2\sigma} \right)
- e^\veps \Phi\left( - \frac{\veps\sigma}{\Delta} - \frac{\Delta}{2\sigma} \right),
\end{equation}
where $\Phi$ denotes the CDF of the standard univariate Gaussian distribution and $\Delta = \norm{d_0 - d_1}_2$. 
\end{lem}

Setting $\alpha=1$ in Eq.~\eqref{eq:alphadiv}, we get the total variation (TV)
distance between the probability distributions $P$ and $Q$ (see, e.g.,~\cite{balle2020privacy}),
\begin{equation} \label{eq:tvdef1}
\begin{aligned}
    \mathrm{TV}(P,Q) &=  \frac{1}{2} \int \abs{P(x) - Q(x)} \, \dd x \\ 
    &= \int [P(x) - Q(x)]_+ \, \dd x.
\end{aligned}
\end{equation}

When $P$ and $Q$ are discrete, defined by probabilities $p_k$ and $q_k$, $k\in \mathbb{Z}$, respectively, we have the important special case of discrete TV distance defined by 
$$
\mathrm{TV}(P,Q) = \sum\nolimits_{k \in \mathbb{Z}} \max \{p_k - q_k,0 \}.
$$

\subsection{Trade-Off Functions and Functional DP} \label{sec:fdp}

DP can also be understood from a hypothesis testing perspective~\cite{wasserman2010statistical}. 
In the context of ML model auditing, this can be formulated as follows~\cite{nasr2023tight}.
Consider the hypothesis testing problem
\begin{equation*}
    \begin{aligned}
&H_0: \quad \textrm{ the model } \theta  \textrm{ is drawn from } P \\
&H_1: \quad \textrm{ the model } \theta  \textrm{ is drawn from } Q,
    \end{aligned}
\end{equation*}
where $P$ and $Q$ are obtained via some post-processing of the probability distributions of $\mathcal{M}(D)$ and $\mathcal{M}(D')$, respectively.
This ensures, in particular, by the post-processing property of DP, that if $\mathcal{M}$ is $(\veps,\delta)$-DP, then $P$ and $Q$ are $(\veps,\delta)$-indistinguishable.

The trade-off function, as defined in~\cite{dong2022gaussian}, captures the difficulty of distinguishing the hypotheses $H_0$ and $H_1$. Given a rejection rule $0 \leq \phi(\theta) \leq 1$ that
takes as an input the model $\theta$ trained by the mechanism $\mathcal{M}$, the type I error is defined as $\alpha_{\phi} = \mathbb{E}_P[\phi]$ and the type II error as $\beta_{\phi} = 1 - \mathbb{E}_Q[\phi]$.
Then, the trade-off function that describes the upper bound for the distinguishability is given as follows.
\begin{defn}
Define the trade-off function $T(P,Q) \, : \, [0,1] \rightarrow [0,1]$ for two probability distributions $P$ and $Q$ as
$$
T(P,Q)(\alpha) = \inf \{ \beta_{\phi} \, : \, \alpha_{\phi} \leq \alpha \}.
$$ 
\end{defn}
For an arbitrary function $f \, : \, [0,1] \rightarrow [0,1]$, the following properties characterize whether it is a trade-off function  (see Prop.~2.2 in~\cite{dong2022gaussian}).
\begin{lem}
A function $f \, : \, [0,1] \rightarrow [0,1]$ is a trade-off function if and only if $f$ is convex, continuous, non-increasing, and $f(x) \leq 1-x$ for all $x \in [0,1]$.    
\end{lem}
The $f$-DP can be then defined as follows.
\begin{defn}
Let $f$ be a trade-off function. A mechanism $\M$ is $f$-DP if
$$
T\big( \M(D), \M(D') \big) \geq f
$$
for all neighboring datasets $D$ and $D'$.
\end{defn}

As shown in~\cite{dong2022gaussian}, $(\veps,\delta)$-DP is equivalent to $f$-DP for the following trade-off function:
$$
f_{\veps,\delta}(\alpha) = \max\{0,1 - \delta - \ee^{\veps} \alpha, \ee^{-\veps}(1 - \delta -  \alpha) \}.
$$
From this, we directly get the following accurate characterization of the trade-off function for a given mechanism $\M$.

\begin{lem} \label{lem:DP_to_T}
 Suppose we have a privacy profile $h(\alpha)$ of the mechanism $\M$. Then, the function given by
$$
f(x) = \max_{ \alpha \geq 0}  \max\{0,1 - h(\alpha) - \alpha x, \alpha^{-1}(1 - h(\alpha) -  x) \}
$$
is a trade-off function of $\M$.   
\end{lem}

This also allows approximating the trade-off function using numerical integrators given a set of points $(\veps_1,\delta_1), \ldots, (\veps_m,\delta_m)$. From Lemma~\ref{lem:DP_to_T} we directly have the following approximation algorithm which is essentially the one given if Appendix A of~\cite{nasr2023tight}.
Notice that having a target delta value $\delta$ ensures that we also obtain accurate $\veps$-values on the interval $[\delta,1-\delta]$ efficiently as the numerical integrators commonly require choosing some interval discretization interval $[-L,L]$ for estimating the privacy loss distributions. Evaluating $\veps$-estimates for delta-values on the interval $[\delta,1-\delta]$ ensures we can use the same approximation of the PLD without incurring additional errors.

\begin{algorithm}[h!]
\caption{Estimation of the trade-off function $f$ using a privacy profile $\delta(\veps)$}
\begin{algorithmic}
\STATE{$F_{\mathcal{M}}$ privacy analysis function that gives that outputs $\veps$ for a given $\delta$, $n$ number of discretization points, $\delta$ target delta in the DP analysis. }
\STATE{$\Delta \leftarrow $ $n$ linearly spaced points on the interval $[\delta,1-\delta]$.}
\FOR{ $\delta' \in \Delta$:}
        \STATE{$\wh{\veps} \leftarrow F_{\mathcal{M}}(\delta')$}
        \STATE{$f_{\delta'}(x) := \max \{0,1-\delta'- x \ee^{\wh{\veps}}, \ee^{- \wh{\veps}}( 1-\delta'- x ) \} $}
\ENDFOR
\STATE{$f(x) := \max_{\delta' \in \Delta} f_{\delta'}(x)$}
\end{algorithmic}
\label{alg:dp_tp_fdp}
\end{algorithm}

Informally speaking, a mechanism is $\mu$-GDP if the outcomes from two neighboring distributions are not more distinguishable than two unit variance Gaussians $\mu$ apart from each other. Using a trade-off function determined by 
$\mathcal{N}(0,1)$ and $\mathcal{N}(\mu,1)$, we have the following charaterization~\cite{dong2022gaussian}.
\begin{defn} \label{def:gdp}
A mechanism $\mathcal{M}$ is $\mu$-GDP if for all $\alpha \in [0,1]$,
$$
T \big(\mathcal{M}(D), \mathcal{M}(D') \big) \geq \Phi ( \Phi^{-1}(1-\alpha) - \mu)
$$
for all neighboring datasets $D,D'$ where $\Phi$ is the standard normal CDF.
\end{defn}

\subsection{Confidence Intervals for $f$-DP}


Using empirical upper bounds for $\alpha$ and $\beta$, obtained using, e.g., the Clopper--Pearson intervals or Jeffreys intervals, and Def.~\ref{def:gdp}, we may obtain an empirical lower bound for the GDP parameter $\mu$ as
\begin{equation}  \label{eq:mu_low}
    \mu_{\mathrm{emp}}^{\mathrm{lower}} = \Phi^{-1}(1-\bar{\alpha}) - \Phi^{-1}(\bar{\beta})
\end{equation}
The work~\cite{nasr2023tight} proposes also to use the \it{credible intervals} \rm for $\veps$ as a basis for the confidence interval estimation in $f$-DP. This approach is based on a certain Bayesian estimation of $\veps$-values proposed in~\cite{zanella2023bayesian}. Therein, given the estimated $\mathrm{FP}$ and $\mathrm{FN}$-values of the attack, a posterior distribution $u_{(\mathrm{FPR},\mathrm{FNR})}(\alpha,\beta)$ is defined as
\begin{equation*}
    \begin{aligned}
        u_{(\mathrm{FPR},\mathrm{FNR})}(\alpha,\beta) = & \mathrm{Beta}(\alpha; 0.5 + \mathrm{FN}, 0.5 + N - \mathrm{FN}) \cdot \\
        & \mathrm{Beta}(\beta; 0.5 + \mathrm{FP}, 0.5 + N - \mathrm{FP})
    \end{aligned}
\end{equation*}
A trade-off curve $f$ is then determined to give an $f$-DP guarantee with confidence $c$, where $c$ is the probability mass of the posterior distribution $u_{(\mathrm{FPR},\mathrm{FNR})}(\alpha,\beta)$ in the privacy region determined by $f$, i.e., between the curves $f(\alpha)$ and $1 - f(1 - \alpha)$, $\alpha \in [0,1]$. 
The confidence value $c$ is then determined by the cumulative distribution function
\begin{equation} \label{eq:Phat}
P_{\hat{.}}(f) = \int\limits_0^1 \int\limits_{f(\alpha)}^{1-f(1-\alpha)} u_{(\mathrm{FPR},\mathrm{FNR})}(\alpha,\beta) \, \dd \beta \dd \alpha
\end{equation}
which gives the mass of the distribution $ u_{(\mathrm{FPR},\mathrm{FNR})}(\alpha,\beta)$ in the privacy region determined by the trade-off function $f$.
Finding a suitable trade-off curve using the integral~\eqref{eq:Phat} is difficult for several reasons and we remark that the work~\cite{nasr2023tight} mostly uses in its experiments the GDP estimate~\ref{eq:mu_low}, where $\bar{\alpha}$ and $\bar{\beta}$ are obtained either using the Clopper--Pearson estimates or Bayesian estimates using the approach of~\cite{zanella2023bayesian}. 

\section{Difficulties in Auditing with $f$-DP}

As shown in~\cite{nasr2023tight}, the success of threshold inference based $\mu$-GDP auditing does not depend on the value of the threshold in case $P \sim \mathcal{N}(1,\sigma^2)$ and $Q \sim \mathcal{N}(0,\sigma^2)$. Asymptotically, we then have that for a threshold value $z \in \mathbb{R}$, $\alpha = 1 - \Phi\left(\tfrac{z}{\sigma} \right)$ and $\beta =  \Phi\left(\tfrac{z-1}{\sigma} \right)$, and for all $z \in \mathbb{R}$,
\begin{equation} \label{eq:mu_est}
\mu = \Phi^{-1}(1-\alpha) - \Phi^{-1}(\beta). 
\end{equation}
However, while the relation~\eqref{eq:mu_est} and the threshold independence hold for $P$ and $Q$ that are exactly Gaussians with an equal variance, they do not hold for general $\mu$-GDP distinguishable distribution and in general finding the GDP parameter accurately requires tuning of the threshold parameter $z$. To illustrate this, consider the example given in~\cite{nasr2023tight}: let $P \sim q \cdot \mathcal{N}(1,\sigma^2) + (1-q) \cdot \mathcal{N}(0,\sigma^2)$ and $Q \sim \cdot \mathcal{N}(0,\sigma^2)$, where $\sigma=0.3$ and $q=0.25$. Using accurate numerical calculation of the privacy profile
$
\delta(\veps) = \max \{ H_{\ee^\veps}\big( P || Q \big), H_{\ee^\veps}\big( Q || P \big) \}
$
and numerical optimization, we find that the pair $(P,Q)$ is $\mu$-GDP for $\mu \approx 1/0.404$ (see Appendix Figure~\ref{fig:0}).

Figure~\ref{fig:1} shows the $\mu$-value estimated using equation~\eqref{eq:mu_est}.
Clearly, the threshold independence of the $\mu$-GDP auditing does not hold for non-Gaussian distributions. Also, we experimentally find that the largest $\mu$-estimate require large threshold values (very small FPRs) so that the confidence intervals easily become large and we are not able to get close to the accurate $\mu$-values even when using $n=10^5$ samples. Also, as we see, finding a suitable value for the threshold value $z$ requires careful tuning as the $\mu$-estimation is $z$-independent only for a pair of Gaussian with equal variance.

\begin{figure}[h!]
\centering
\includegraphics[width=0.6\columnwidth]{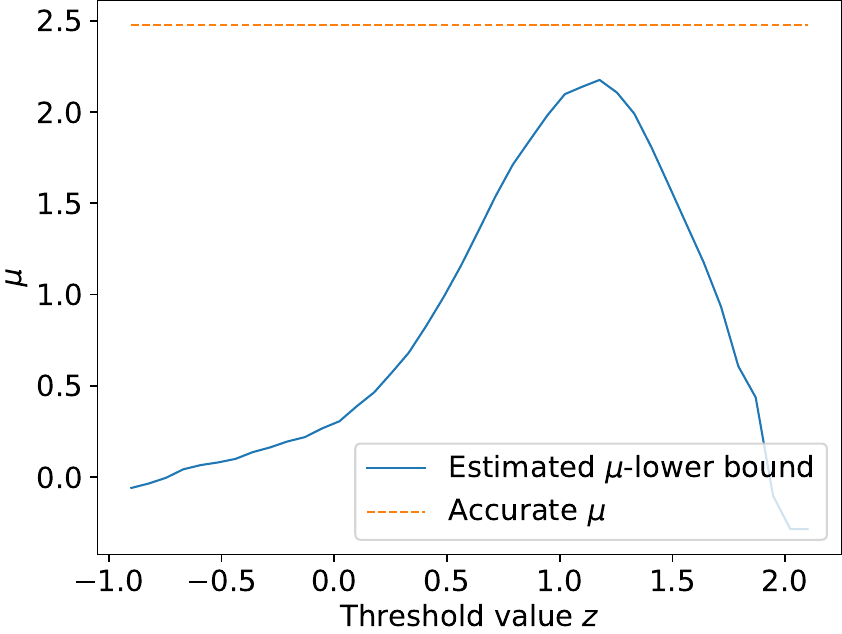} 
\caption{Adjusting the $\mu$-GDP parameter for the pair of distributions $P \sim q \cdot \mathcal{N}(1,\sigma^2) + (1-q) \cdot \mathcal{N}(0,\sigma^2)$ and $Q \sim \cdot \mathcal{N}(0,\sigma^2)$ using a threshold attack with threshold value $z \in \mathbb{R}$. The figure shows the estimated $\mu$-value as a function of $z$. Each $\mu$-lower bound value is estimated using the Clopper--Pearson confidence intervals and $n=10^4$ samples from both $P$ and $Q$. 
}
\label{fig:1}
\end{figure}



As we show in the next section, we can relax the requirement of knowing any parameters or even any parametric form of the distributions $P$ and $Q$ and still be able to accurately audit the subsampled Gaussian mechanism and many others.

In the general case, such as when carrying our $f$-DP auditing of the subsampled Gaussian mechanism, one has to use the formula~\eqref{eq:Phat}.
The first major difficulty with using the integral~\eqref{eq:Phat} one encounters is in the case when auditing mechanisms determined by more than one parameter. For example, when auditing the subsampled Gaussian mechanism, the potential $f$-curves are parameterized by two parameters, $q$ and $\sigma$. Thus, given only the observations, it is not obvious how to adapt $q$ and $\sigma$ to obtain high-confidence privacy regions for the posterior distribution $u_{(\mathrm{FPR},\mathrm{FNR})}(\alpha,\beta)$ as both of these parameters will affect the shape of the trade-off function $f$. If one is focused on point-wise $(\veps,\delta)$-DP estimates one may end up with wildly different $f$-DP guarantees: as demonstrated recently in~\cite{kaissisbeyond}, two mechanisms can have wildly different privacy profiles while having the same point-wise $(\veps,\delta)$-DP guarantees.

The second difficulty one quickly encounters with the formula~\eqref{eq:Phat} is the numerical approximation. The formula~\eqref{eq:Phat} which does not seem to exhibit analytical solutions even in the simplest cases (e.g., $\mu$-GDP estimation). Therein, one specific issue that requires careful attention is that even the $f(\alpha)$-curve that determines the boundary of the privacy region may not have analytical expression but has to be approximated numerically. This is the case, e.g., in case $f$ is a trade-off curve of the subsampled Gaussian mechanism, where we approximate it using Algorithm~\ref{alg:dp_tp_fdp}. However, the biggest difficulty seems to arise from the numerical stability of the integration. 



\begin{figure}[h!]
\centering
\includegraphics[width=0.6\columnwidth]{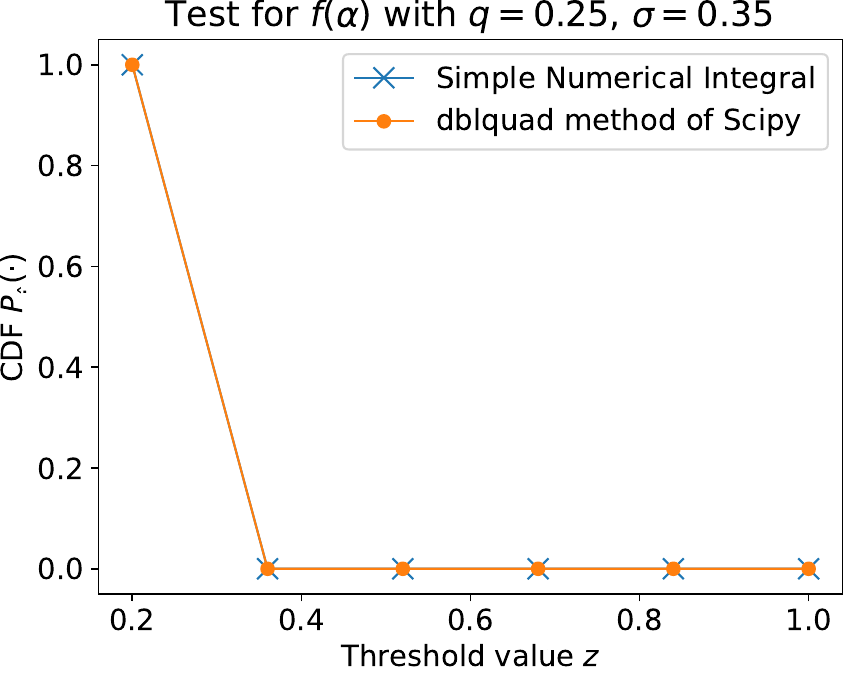} 
\caption{Estimate of the cumulative density function $P_{\hat{.}}(\cdot)$, i.e., the probability mass of the posterior distribution $u_{(\mathrm{FPR},\mathrm{FNR})}(\alpha,\beta)$ inside the privacy region determined by the trade-off function $f$ of the subsampled Gaussian mechanism with sampling ratio $q=0.25$ and noise parameter $\sigma=0.35$. Using a threshold value between -0.75 and 0.2 would lead us to conclude with high confidence that the mass of $u_{(\mathrm{FPR},\mathrm{FNR})}(\alpha,\beta)$ is inside the privacy region. While this would lead to a correct lower bound for the DP parameter $\veps$, it would give an inaccurate approximation of the true trade-off function.
}
\label{fig:x1}
\end{figure}

\begin{figure}[h!]
\centering
\includegraphics[width=0.6\columnwidth]{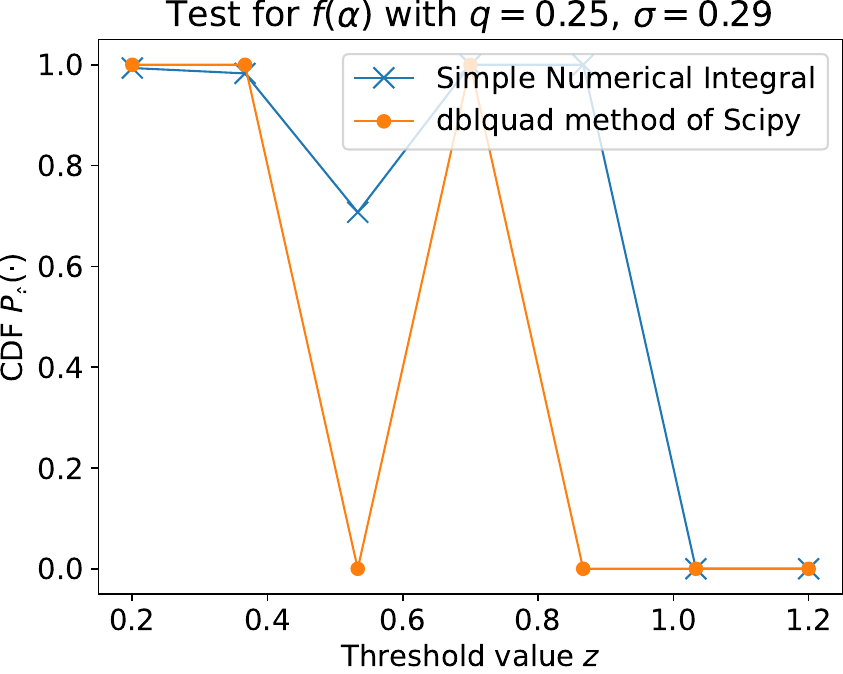} 
\caption{Estimate of the cumulative density function $P_{\hat{.}}(\cdot)$, i.e., the probability mass of the posterior distribution $u_{(\mathrm{FPR},\mathrm{FNR})}(\alpha,\beta)$ inside the privacy region determined by the trade-off function $f$ of the subsampled Gaussian mechanism with sampling ratio $q=0.25$ and noise parameter $\sigma=0.35$. Using a threshold value between -0.75 and 0.2 would lead us to conclude with high confidence that the mass of $u_{(\mathrm{FPR},\mathrm{FNR})}(\alpha,\beta)$ is inside the privacy region. While this would lead to a correct lower bound for the DP parameter $\veps$, it would give an inaccurate approximation of the true trade-off function.
}
\label{fig:x2}
\end{figure}

We demonstrate the difficulty of the numerical $f$-DP auditing with an example where we are auditing the one-dimensional distributions $P \sim q \cdot \mathcal{N}(1,\sigma^2) + (1-q) \cdot \mathcal{N}(0,\sigma^2)$ and $Q \sim \cdot \mathcal{N}(0,\sigma^2)$ where $q=0.25$ and $\sigma=0.3$. We consider a situation where the auditor is given the value of $q$ and is trying to determine the upper bound $f$-trade-off curve by scaling $\sigma$ and by using a numerical approximation of the integral~\eqref{eq:Phat}. The posterior distribution $u_{(\mathrm{FPR},\mathrm{FNR})}$ is constructed using threshold membership inference and $n=10^5$ samples from both $P$ and $Q$. To reduce the influence of the numerical integrator on our conclusions, we use two different numerical integration methods:
we use the $\mathrm{dblquad}$-integrator included in $\mathrm{scipy.integrate}$ library~\cite{virtanen2020fundamental} and a simple two-dimensional Euler method.
For a given value of $\sigma$, we compute an approximation of the accurate $f$-DP curve of the subsampled Gaussian mechanism using Alg.~\ref{alg:dp_tp_fdp}with 200 points and estimate the true $f$-function by a piece-wise linear function constructed using these points. The privacy region estimated using this approximated $f$-curve is given as an integral region for the $\mathrm{dblquad}$-integrator.
When $\sigma=0.35$, both integrators correctly indicate that the $f$-curve is an upper bound for the privacy region (Fig.~\ref{fig:x1}). However, when $\sigma=0.29$, we can still find threshold values for which both integrators would deem the privacy region to be under the $f$-curve, which is clearly a wrong conclusion (Fig.~\ref{fig:x2}).

\section{Histogram-Based Auditing of DP Guarantees} \label{sec:scores}

We next present our histogram-based DP auditing method that does not require any a priori information about the underlying DP mechanism.
\subsection{Problem Formulation}

Similarly to the hypothesis testing formulation of DP presented in Section~\ref{sec:dp}, our method is based on a general problem formulation, where the privacy profile of the underlying DP mechanism $\M$ dominates the privacy profile $h(\alpha) = H_{\alpha}(P,Q)$ determined by some distributions $P$ and $Q$ and we have a number of independent samples from both $P$ and $Q$. 
Then, having an estimate (or high-confidence lower bound) for $h(\alpha)$ will also give a lower bound for the privacy profile of $\M$.

We can motivate this formulation for example via black-box auditing of an ML model training algorithm $\M$ as follows. 
Let $\theta \in \mathbb{R}^d$ denote the ML model parameters, $F(\theta,x)$ the forward mapping for the feature $x$ of a data element $z = (x,y)$, where $y$ denotes the label, and let $\ell\big(F(\theta,z),y \big)$ be some loss function. Then, by the post-processing property of DP, the distributions 
$$
P = \ell\big(F(\theta,x),y \big), \quad \theta \sim \M(D),
$$
and 
$$
Q = \ell\big(F(\theta,x),y \big), \quad \theta \sim \M(D \cup z)
$$
are $(\veps,\delta)$-close to each other, i.e., $H_{\ee^{\veps}}(P,Q) \leq \delta$.

We next show how to lower bound the privacy profile $h(\alpha)$ using histogram density estimates of the distributions $P$ and $Q$.


\subsection{Estimating Hockey-Stick Divergence Using Histograms} \label{sec:est_hist}

We estimate the distributions $P$ and $Q$ by first sampling $n$ samples from $P$ and $n$ samples from $Q$, and then 
using binning such that we place the score  values into $k$ bins, each of given width $h>0$.
Denote these samples by $P_S = \{P_1,\ldots,P_{n} \}$ and $Q_S = \{Q_1,\ldots,Q_{n} \}$
Notice that we could use an adaptive division of the real line to generate the bin, however we here focus
on equidistant bins for simplicity. Also, we could consider drawing a different amount of samples from $P$ and $Q$.
Given left and right end points $a$ and $b$, respectively, we define the bin $j$, $j \in \{2,\ldots,k-1\}$, as
$$
\textrm{Bin}_j = [a + (j-1)\cdot h, a + j \cdot h)
$$
and
$$
\textrm{Bin}_1 = (\infty, a+h), \quad \textrm{Bin}_k = [b-h, \infty).
$$
We define the probabilities $p_j$ and $q_j$, $j \in [k]$, by the relative frequencies of $P$'s and $Q$'s samples hitting bin $j$:
\begin{equation*}
    \begin{aligned}
        p_j & \leftarrow \frac{1}{n}  \abs{ \{ x \in P_S \, : \, x \in \textrm{Bin}_j \} }, \\
q_j & \leftarrow \frac{1}{n}  \abs{ \{ x \in Q_S \, : \, x \in \textrm{Bin}_j \} }.
    \end{aligned}
\end{equation*}
Denote these estimated discrete distributions with probabilities $p_j$ and $q_j$, $j \in [k]$, by $\wh P$ and $\wh Q$, respectively. Then, we estimate the  parameters of the mechanism $\mathcal{M}$ by using the hockey-stick divergence $H_{\ee^{\veps}}(\wh P||\wh Q)$, $\veps \in \mathbb{R}$. This is motivated by the following observation.

\begin{lem}
    Denote the limiting distributions by $\wt P$ and $\wt Q$, i.e., 
$$
\wt P_j = \int_{\textrm{Bin}_j} P(t) \, \dd t, \quad \textrm{and} \quad 
\wt Q_j = \int_{\textrm{Bin}_j} Q(t) \, \dd t
$$
for all $j \in [k]$, where $P(t)$ and $Q(t)$ denote the density functions of $P$ and $Q$, respectively.
Then, for all $\veps \in \mathbb{R}$:
$$
H_{\ee^{\veps}}(\wt P,\wt Q) \leq H_{\ee^{\veps}}(P,Q).
$$
\begin{proof}
The distributions  $\wt P$ and $\wt Q$ are obtained by applying the same post-processing function to $P$ and $Q$ and the claim follows from the data processing inequality.
\end{proof}
\end{lem}

To obtain a high-confidence lower bound for the hockey-stick divergence $H_{\ee^{\veps}}(\wt P,\wt Q)$,
the challenge is then how bound the error in the estimate $H_{\ee^{\veps}}(\wh P||\wh Q)$.
We next show how to obtain frequentist confidence intervals for this estimate.

\subsection{Confidence Intervals for Histogram-Based $\veps$-Estimates}

We consider frequentist confidence intervals, and thus by definition a $(1-\alpha)$-confidence interval will contain the true parameter with $(1-\alpha)$\% of the time the estimation is carried out.


An important observation here is that the counts of samples hitting bins, 
$$
\abs{\{ x \in P_S \, : \, x \in \textrm{Bin}_j \}} \quad \textrm{and} \quad
\abs{ \{ x \in Q_S \, : \, x \in \textrm{Bin}_j \} },
$$ 
$j \in [k]$, are independent draws from multinomial distributions with $k$ events and event probabilities $\wt P$ and $\wt Q$, respectively.
Denote the set of possible multinomial probabilities for the discrete set $X$ by
$$
\Delta(X) = \{ p \in \mathbb{R}_{\geq 0}^{\abs{X}} \, : \, \norm{p}_1 = 1 \}.
$$
To obtain confidence intervals, we use the following high-probability bound for the total variation distance given in~\cite{canonne2020short}.

\begin{lem} \label{lem:canonne}
Consider the empirical distribution $\wt p$ obtained by drawing $n$ independent samples $s_1, \ldots, s_n$ from the underlying distribution $p \in \Delta([k])$:
$$
\wt p_i = \frac{1}{n} \abs{ \{ s \in \{s_1, \ldots, s_n \} \, : \, s = i  \} }, \quad i \in [k].
$$
Then, as long as 
$$
n \geq \max \left\{ \frac{k}{\veps^2}, \frac{2}{\veps^2} \log \frac{2}{\delta} \right\},
$$
we have that with probability at least $1-\delta$, 
$$
\mathrm{TV}(p,\wt p) \leq \veps.
$$
\end{lem}

It is evident that by choosing $n$ as guided by Lemma~\ref{lem:canonne}, the interval $[\wt p - \veps, \wt p + \veps]$
will be a $100 \% \cdot (1-\delta)$ - confidence interval for the TV distance estimate.
We can use the confidence interval for TV distance also to obtain high-confidence lower bounds for other parts of the privacy profile via the following result.

\begin{lem} \label{lem:TV_to_eps_delta_bound}
Denote $P,Q$ probability distributions on the same probability space. Suppose 
$$
\mathrm{TV}(P,\wt P) \leq \tau
$$
and
$$
\mathrm{TV}(Q,\wt Q) \leq \tau
$$
for some $\tau \geq 0$. Then, for all $\veps \in \mathbb{R}$,
$$
H_{\ee^{\veps}}(P,Q) \leq H_{\ee^{\veps}}(\wt P, \wt Q) + (1+\ee^{\veps}) \cdot \tau.
$$
    
\end{lem}
For obtaining high-confidence $f$-DP upper bounds, our strategy is to determine the high-confidence lower bounds for the privacy profile using Lemma~\ref{lem:TV_to_eps_delta_bound} and then convert these lower bounds to trade-off functions using Lemma~\ref{lem:DP_to_T}. We remark that rigorously, this approach does not give a high-confidence $f$-DP upper bound. Due to the convexity of the trade-off functions, using point-wise upper bounds for the privacy profile would give a high-confidence lower bound for the trade-off function, however for the upper bound we would need to use an approach similar to that of~\cite{doroshenko2022connect}, where they give an optimistic numerical approximation of the privacy profile that strictly lower bounds the true privacy profile. We believe however that the effect would be small and in experiments we simply use as a high-confidence upper bound the trade-off function approximated using Lemma~\ref{lem:DP_to_T}.

\subsection{Convergence Result for the Hockey-Stick Divergence Estimate}

The density estimation using histograms is a classical problem in statistics, and existing results such as those of~\cite{scott1979optimal} can be used to derive suitable bin widths for the histograms. We also mention the work~\cite{wand1997data} which gives methods based on kernel estimation theory and the work~\cite{knuth2013optimal}  which gives binning based on a Bayesian procedure.

Consider the approach and notation of Section~\ref{sec:est_hist}, except that for the theoretical analysis we consider an infinite number of bins and focus on find the optimal bin width $h$. I.e., we define the bins  such that for $j \in \mathbb{Z}$,
$$
\textrm{Bin}_j = [j \cdot h, (j+1) \cdot h),
$$
and place the $n$ randomly drawn samples from $P$ and $Q$ into these bins to estimate the probabilities 
$\int_{\textrm{Bin}_j} P(x) \dd x$ and $\int_{\textrm{Bin}_j} Q(x) \dd x$
using the bin-wise frequencies of the histograms.
If we denote the piece-wise continuous density function as
$$
\wh P(x) = \wh P_j/h, \quad \textrm{when} \quad x \in \textrm{Bin}_j,
$$ 
then the analysis of~\cite{scott1979optimal} gives an optimal bin width for minimizing the mean-square error $\mathbb{E}(P(x) - \wh P(x))^2$ for a density function $P(x)$ with bounded and continuous derivatives up to second order (and similarly for $Q$).  We can directly use this result for analysing the convergence of the numerical hockey-stick divergence $H_{\ee^{\veps}}(\wh P||\wh  Q)$, $\veps \in \mathbb{R}$, as a function of the number of samples $n$.

\begin{thm} \label{thm:bin_convergence}
Let $P$ and $Q$ be one-dimensional probability distributions with differentiable density functions $P(x)$ and $Q(x)$, respectively, and
consider the histogram-based density estimation described above. Draw $n$ samples both from $P$ and $Q$, giving density estimators $\wh P = (\wh P_1, \ldots, \wh P_k)$ and $\wh Q = (\wh Q_1, \ldots, \wh Q_k)$, respectively.
Let the bin width be chosen as
\begin{equation} \label{eq:optimal_h}
h_n = \left( \frac{12}{\int P'(x)^2 \, \dd x + \int Q'(x)^2 \, \dd x} \right)^{\frac{1}{3}} n^{-\frac{1}{3}}
\end{equation}
Then, for any $\alpha \geq 0$, the numerical hockey-stick divergence $H_\alpha(\wh P || \wh Q)$ convergences in expectation to 
$H_\alpha(P || Q)$ with rate $\mathcal{O}(n^{-1/3})$, i.e.,
$$
\mathbb{E}\abs{H_\alpha(\wh P || \wh Q) - H_\alpha(P || Q) } = \mathcal{O}(n^{-1/3}),
$$
where the expectation is taken over the random draws for constructing $\wh P$ and $\wh Q$.
\end{thm}
In case $P$ and $Q$ are Gaussians with an equal variance, we directly get the following the analysis of Section 3 of~\cite{scott1979optimal}.
\begin{cor}
Suppose $P$ and $Q$ are one-dimensional normal distributions both with variance $\sigma^2$. Then, the bin width $h_k$ of Eq.~\eqref{eq:optimal_h} is given by
\begin{equation} \label{eq:bin_width_gauss}
h_n = 2 \cdot 3^{1/3} \cdot \pi^{1/6} \cdot \sigma \cdot n^{-1/3}.
\end{equation}
\end{cor}
We may use the expression of Eq.~\eqref{eq:bin_width_gauss} for Gaussians as a rule of thumb also for other distributions with $\sigma$ denoting the standard deviation.


\subsection{Pseudocode for the Histogram-Based Estimation of DP-Guarantees}

The pseudocode for our $(\veps,\delta)$-DP auditing method is given in Alg.~\ref{alg:auditing_method}. 
Notice that in  in order to find a suitable bin width $h$, we may also estimate the standard deviations of the samples $P_S$ and $Q_S$. This is also motivated by the experimental observation that the variances of the score values for auditing training and test sets are similar. Then, having an std estimate $\sigma$, we could set the bin width $h=3.5 \cdot n^{-1/3} \wh \sigma$ which approximately equals the expression~\eqref{eq:bin_width_gauss}.

\begin{algorithm}[h!]
\caption{Estimation of $(\veps,\delta)$-DP parameters Using Histogram Density Estimation}
\begin{algorithmic}
\STATE{\textbf{Input:} $n$ indepedent samples from the distributions $P$ and $Q$: $P_S = \{P_1,\ldots,P_n \}$ and $Q_S = \{Q_1,\ldots,Q_n \}$, DP parameter $\delta \in (0,1)$. Number of Bins $k$, end points $a,b \in \mathbb{R}$.}
\STATE{Set the bin width $h=\frac{b-a}{k}$.}
\STATE{Divide the real line into $k$ disjoint intervals such that for  $j \in \{2,\ldots,k-1\}$,
$$
\textrm{Bin}_j = [a + (j-1)\cdot h, a + j \cdot h)
$$
and $\textrm{Bin}_1 = (\infty, a+h)$ and $\textrm{Bin}_k = [b-h, \infty).$
}
\STATE{
Estimate the probabilities $p_j$ and $q_j$, $j \in [k]$, by the relative frequencies of hitting bin $j$ as
\begin{equation*}
    \begin{aligned}
        p_j & \leftarrow \frac{1}{n}  \abs{ \{ x \in P_S \, : \, x \in \textrm{Bin}_j \} }, \\
q_j & \leftarrow \frac{1}{n}  \abs{ \{ x \in Q_S \, : \, x \in \textrm{Bin}_j \} }
    \end{aligned}
\end{equation*}
giving the discrete-valued  distributions \\
$\wh P = \{p_i\}_{i=1}^k$ and $\wh Q = \{q_i\}_{i=1}^k.$ }
\STATE{Set: $\delta \leftarrow H_{\ee^{\veps}}\big(\wh P || \wh Q \big).$}
\RETURN{$\delta$.}
\end{algorithmic}
\label{alg:auditing_method}
\end{algorithm}

 \section{Numerical Examples} \label{sec:num_examples}

Next, we give numerical examples to illustrate the histogram-based estimation presented in Section~\ref{sec:scores}.

\subsection{Numerical Example: Estimating TV Distance Between Two Gaussians}

\begin{figure}[h!]
\centering
\includegraphics[width=0.6\columnwidth]{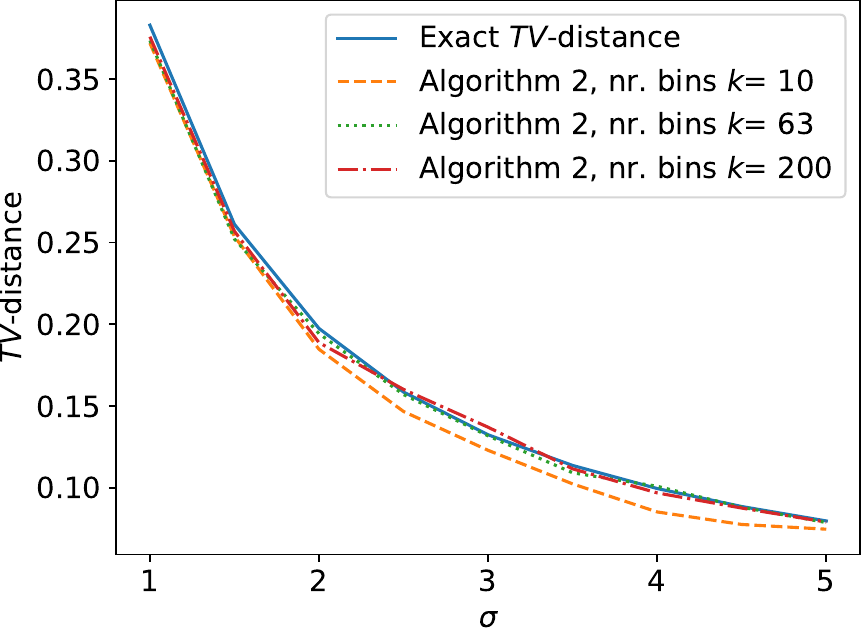}
\caption{Exact TV distance $TV(P,Q)$ and the TV distance approximated using Alg.~\ref{alg:auditing_method} for different values of $\sigma$, when $k=50000$.
The bin width $h_n$ set using Eq.~\eqref{eq:bin_width_gauss} gives $k=63$ bins.}
\label{fig:TV_Gaussx}
\end{figure}

We illustrate our approach for estimating the $(\veps,\delta)$-distance between two one-dimensional Gaussians. The example also illustrates the effect of the  bin size.  Let $\sigma > 0$. We draw $n$ random samples $x_1, \ldots, x_n$ from the distribution $P \sim \mathcal{N}(0,\sigma^2)$ and $n$ samples $y_1, \ldots, y_n$ from the distribution $Q \sim \mathcal{N}(z,\sigma^2).$
We know that $P$ and $Q$ are $\big(\veps,\delta(\veps) \big)$-distinguishable, where $\delta(\veps)$
denotes the privacy profile of the Gaussian mechanism with noise scale $\sigma$ and sensitivity 1
and in particular 
%
we know by Lemma~\ref{lem:gauss_dp} that the total variation distance $\mathrm{TV}(P,Q)$ is given by 
\begin{equation} \label{eq:TV_gaussian_x}
\begin{aligned}
     \delta(0)
= 2 \cdot \left( 1 - \Phi\left( \frac{1}{2\sigma} \right) \right),
\end{aligned}
\end{equation}
where $\Phi$ denotes the CDF of the standard univariate Gaussian distribution.
We determine $a$ and $b$ such that $x_i$'s and $y_i$'s are inside the interval $[a,b]$ with high probability and fix the number of bins $N \in \mathbb{N}$, and carry out the TV distance estimation using Algorithm~\ref{alg:auditing_method} (i.e., using $\veps=0$). 
Figure~\ref{fig:TV_Gaussx} illustrates the accuracy of the TV distance estimation as the number of bins $N$ varies.

\subsection{Numerical Example: Auditing the Subsampled Gaussian Mechanism}

\begin{figure}[h!]
\centering
\includegraphics[width=0.6\columnwidth]{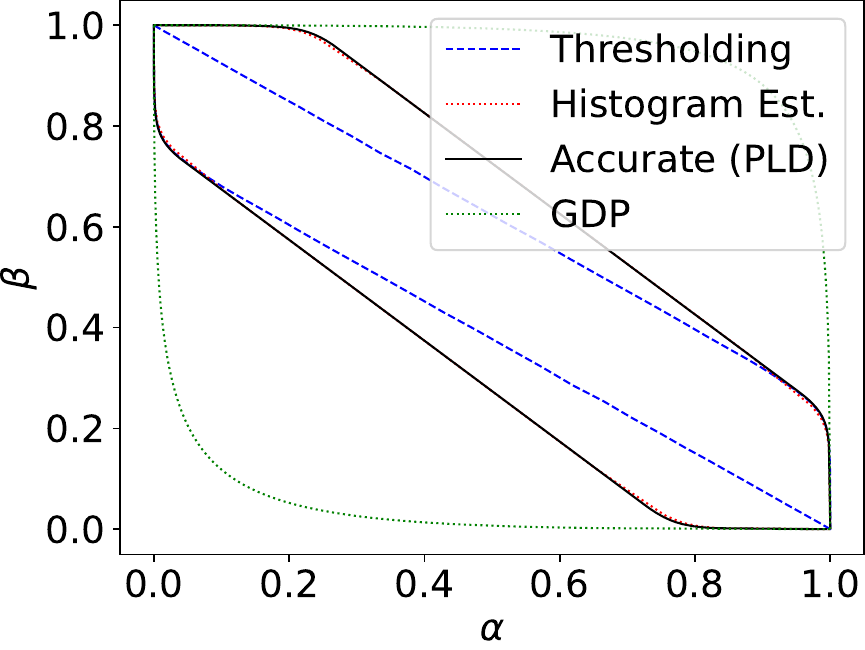}
\caption{Estimating the trade-off function of the subsampled Gaussian mechanism with $q=\tfrac{1}{4}$ and $\sigma=0.3$. The histogram-based auditing method is able to accurately estimate the trade-off function without any information about $P$ and $Q$. We sample $n=10^5$ samples from both $P$ and $Q$.}
\label{fig:fnrfpr}
\end{figure}

In~\cite{nasr2023tight} an open problem of how to accurately audit the subsampled Gaussian mechanism is posed. The concrete example of~\cite{nasr2023tight} considers the pair of distributions $P \sim q \cdot \mathcal{N}(1,\sigma^2) + (1-q) \cdot \mathcal{N}(0,\sigma^2)$ and $Q \sim \mathcal{N}(0,\sigma^2)$ with the parameter values $q=1/4$ and $\sigma=0.3$. Figure~\ref{fig:fnrfpr} replicates the experimental results given in~\cite{nasr2023tight}, however, it includes the trade-off function estimated using Alg.~\ref{alg:auditing_method}. The accurate trade-off curve is computed using numerical privacy accounting method of~\cite{koskela2021tight} and Alg.~\ref{alg:dp_tp_fdp}. We see that the histogram-based method is able to accurately estimate this trade-off curve.

\subsection{Numerical Example: Auditing the Laplace Mechanism}

The Laplace mechanism adds Laplace distributed noise to a function with limited  $L_1$-sensitivity, and the $(\veps,\delta)$-DP privacy guarantees are determined by a dominating pair of distributions $P\sim \mathrm{Lap}(0,\lambda)$ and $Q\sim \mathrm{Lap}(\Delta_1,\lambda)$, where $\Delta_1$ is the $L_1$-norm sensitivity of the underlying function and $\lambda$ denotes the noise scale. In~\cite{dong2022gaussian} it is shows that the accurate trade-off function of the Laplace mechanism is given by
\begin{equation*}
             T \big( \mathrm{Lap}(0,\lambda),\mathrm{Lap}(\Delta_1,\lambda) \big)(\alpha) = F\big( F^{-1}(1-\alpha) - \mu \big)
\end{equation*}
where $\mu = \lambda/\Delta_1$ and $F$ denotes the CDF of $\mathrm{Lap}(0,1)$ (see Appendix for the exact analytical form).

Figure~\ref{fig:fnrfpr_laplace} shows that the binning-based method is able to accurately estimate the exact trade-off function without any information about the underlying distributions. To compute the trade-off functions, we use $k=10^5$ samples and $100$ bins.

\begin{figure}[h!]
\centering
\includegraphics[width=0.6\columnwidth]{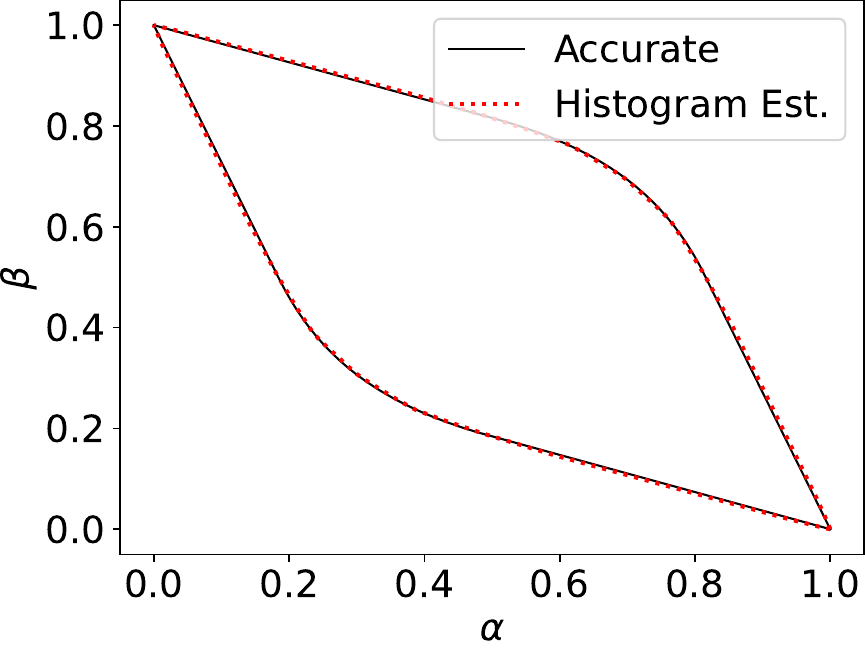}
\caption{
 Estimation of the trade-off function of the Laplace mechanism with noise scale $\lambda=1.0$ and sensitivity $\Delta_1=1.0$. The histogram-based auditing method is able to accurately estimate the trade-off function without any information about $P$ and $Q$. We sample $n=10^5$ samples from both $P$ and $Q$.}
\label{fig:fnrfpr_laplace}
\end{figure}

\section{Relation to Existing Work on Threshold Membership Auditing} \label{sec:relation}

As we next show, the commonly considered membership inference attacks can be seen as a special case of our auditing method.
Suppose that we are given some a fixed threshold $\tau$. We infer that a sample is in the auditing training set in case its score is below $\tau$. This gives the true positive ratios (TPRs) and false positive ratios (FPRs)
\begin{equation} \label{eq:TPRFPR}
    \begin{aligned}
        \mathrm{TPR} &= \mathbb{P}_{x \sim P} \left( S(\theta,x) < \tau  \right), \\
        \mathrm{FPR} &= \mathbb{P}_{x \sim Q} \left( S(\theta,x) < \tau  \right).
    \end{aligned}
\end{equation}
We can interpret the $(\veps,\delta)$-estimates given by this threshold membership inference as the $(\veps,\delta)$-distance between two-bin approximations (bins defined by the parameter $\tau \in \mathbb{R}$ dividing the real line into two bins) of the distributions $P$ and $Q$.

Let $P_2$ and $Q_2$ denote the two-bin histogram approximations of $P$ and $Q$, respectively, where the bins are determined by the threshold parameter $\tau$.
The following lemma shows that the $(\veps,\delta)$-distance between $P_2$ and $Q_2$ exactly matches with the expression commonly used for the empirical $\veps$-values.


\begin{lem} \label{lem:kairouz}
Consider two probability distributions $P$ and $Q$ and distributions $P_2$ and $Q_2$ obtained with two-bin frequency histograms defined by a threshold $\tau \in \mathbb{R}$. Suppose the underlying mechanism $\mathcal{M}$ is $(\veps,\delta)$-DP for some $\veps \geq 0$ and $\delta \in [0,1]$. Then, asymptotically,
$$
\max\{ H_{\ee^{\veps}}\big( P_2 || Q_2 \big), H_{\ee^{\veps}}\big( Q_2 || P_2 \big) \} \leq \delta,
$$
where
\begin{equation} \label{eq:kairouz_c}
        \veps = \max \left\{ \log \frac{\mathrm{TPR} - \delta}{\mathrm{FPR}}, \log \frac{\mathrm{TNR} - \delta}{\mathrm{FNR}} \right\}.
\end{equation}
and $\mathrm{TPR}$ and $\mathrm{FPR}$ are as defined in Eq.~\eqref{eq:TPRFPR} and $\mathrm{FNR} = 1 - \mathrm{TPR}$ and $\mathrm{TNR} = 1 - \mathrm{FPR}$.
\end{lem}
The $\veps$-estimate of Eq.\eqref{eq:kairouz_c} is the commonly used characterization for the connection between the success rates of membership inference attacks and the $(\veps,\delta)$-DP guarantees of the underlying mechanism. Our novelty is to generalize the auditing based on Eq.\eqref{eq:kairouz_c} such that we consider histograms with more than two bins, and instead of estimating TPRs and FPRs, we estimate the relative frequencies of the scores hitting each of the bins and then measure the $(\veps,\delta)$-distance between the approximated distributions corresponding to the score distributions of the two auditing sets.

As an example, suppose that we have a division  of an interval into $2^k$ bins, $k\in \mathbb{N}$, denoted $D_k$, such that half of the bins are right to the threshold $\tau$ and half of them are left to $\tau$ and suppose the division $D_{k+1}$ is obtained by dividing each interval of $D_k$ in half. Then, by the post-processing property, the asymptotic distributions $P_k$ and $Q_k$ obtained using the histogram $D_k$ can be seen as a post-processing of the distributions $P_{k+1}$ and $Q_{k+1}$ (simply sum up the probabilities of adjacent bins) and therefore the finer the division the closer the $(\veps,\delta)$-estimates get to the actual $(\veps,\delta)$-distance between the distributions of the scores.

Following the discussion of~\cite{jagielski2023note}, we see that 
our approach is also related to the exposure metric defined in~\cite{carlini2019secret}. Given $n$ auditing training samples $\{c_i\}_{i=1}^n$ and $n$ auditing test samples $\{r_i\}_{i=1}^n$,~\cite{carlini2019secret} defines the exposure of a sample $c_i$ via its rank
$$
\mathrm{Exposure}(c_i) = \log_2 n - \log_2 \mathrm{rank}(c_i, \{r_i\}_{i=1}^n),
$$
where $\mathrm{rank}(c_i, \{r_i\}_{i=1}^n)$ equals the number of auditing test samples with loss smaller then the loss of $c_i$. As shown in~\cite{jagielski2023note}, a reasonable approximation for the the expected exposure is given by the threshold membership inference (i.e., a two-bin histogram approximation desribed above) with threshold parameter $\tau = \ell_{\mathrm{median}}$, where $\ell_{\mathrm{median}}$ is the median value of the losses of the auditing training samples, i.e., the median of  $\{\ell(c_i)\}_{i=1}^n$. This leads to the $\veps$-estimate given by Eq.~\eqref{eq:kairouz_c} with
$$
\mathrm{TPR} = \mathbb{P}_{x \sim \{c_i\}_{i=1}^n} \left( \ell(x) < \ell_{\mathrm{median}}  \right)
$$ 
and
$$
\mathrm{FPR} = \mathbb{P}_{x \sim \{r_i\}_{i=1}^n} \left( \ell(x) < \ell_{\mathrm{median}}  \right).
$$

We remark that the auditing training and test sample scores would generally need to be independent to conclude that the estimate of Eq.\eqref{eq:kairouz_c} gives a lower bound for the actual $(\veps,\delta)$-DP guarantees.



\section{Lower Bound for a Single Parameter Using TV Distance} \label{sec:tv}

We could in principle use any hockey-stick divergence to estimate the privacy profile
of a mechanism $\mathcal{M}$ in case we can parameterize the privacy profile with a single real-valued parameter in a way that the privacy guarantees depend monotonically on that parameter. Consider, for example, the noise level $\sigma$ for the Gaussian mechanism with sensitivity 1, where finding the $\delta$-value for any $\veps \in \mathbb{R}$ will also give a unique value for $\sigma$.
This kind of single-parameter dependence serves as a good heuristics for analyzing DP-SGD trained models, as the privacy profiles for large compositions are commonly very close to those of a Gaussian mechanism with a given noise scale~\cite{dong2022gaussian}.

Thus, given an estimate of any hockey-stick divergence between the frequency estimates $\wh P$ and $\wh Q$ for an DP-SGD trained model, we get an estimate of the whole privacy profile and in particular get an estimate of an $\veps$-value for a fixed $\delta$-value. 
Figure~\ref{fig:2} (Appendix) illustrates this by showing the relationship between the TV distances  and $\veps$-values for a fixed $\delta>0$ for the Gaussian mechanism, obtained by varying the noise parameter $\sigma$. I.e., the parameter $\sigma$ is first numerically determined using the TV distance and the analytical expression of Eq.~\ref{eq:TV_gaussian_x}, and then the $\veps$-value is numerically determined using the analytical expression of Eq.~\eqref{eq:delta_gaussian}.

We next analytically show that the choice $\alpha=1$, i.e., the TV distance, in fact gives an estimator that is not far from optimal among all hockey-stick divergences for estimating the distance between two Gaussians.

\subsection{Optimal Choice of Hockey-Stick: Total Variation Distance}

In principle, we could use any hockey-stick divergence to estimate the statistical distance between the frequency estimates $\wh P$ and $\wh Q$ and to subsequently deduce the parameter of the underlying mechanism $\mathcal{M}$. However, experiments indicate that the TV distance is generally not far from optimum for this procedure. This is analytically explained by the following example.

Consider two one-dimensional Gaussians $P_\sigma \sim \mathcal{N}(0,\sigma^2)$ and $Q_\sigma \sim \mathcal{N}(1,\sigma^2)$. We first rigorously show that there is a one-to-one relationship between the hockey-stick divergence values and $\sigma$, i.e., that the hockey-stick divergence $H_\alpha(P_\sigma||Q_\sigma)$ is an invertible function of $\sigma$ for all $\sigma \in (0,\infty)$ for all $\alpha > 0$.
\begin{lem} \label{lem:invertibility}
Let $\alpha > 0$. The hockey-stick divergence $H_\alpha\big(P_\sigma||Q_\sigma\big)$ as a function of $\sigma$ is invertible for all $\sigma > 0$.
\end{lem}
Denote $F_\alpha(\sigma) := H_\alpha(P_\sigma || Q_\sigma)$.
To find a robust estimator, we would like to find an order $\alpha > 0$ such that the $\sigma$-value that we obtain using the numerical approach would be least sensitive to errors in the evaluated $\alpha$-divergence.
If we have an error $\approx \Delta H$ in the estimated $\alpha$-divergence, we would approximately have an error $\Delta \sigma = \abs{ \frac{\dd}{\dd H} \, F^{-1}_\alpha (H) } \cdot \Delta H$ in the estimated $\sigma$-value. Thus, we want to solve
$$
\mathrm{arg min}_{\alpha > 0} \, \abs{ \frac{\dd}{\dd H} \, F^{-1}_\alpha (H) }. 
$$
By the inverse function rule, if $H = F_{\alpha}(\sigma)$, we have that
\begin{equation} \label{eq:inverse_fct_rule}
\begin{aligned}
    \mathrm{arg min}_{\alpha > 0} \, \abs{ \frac{\dd}{\dd H} \, F^{-1}_\alpha (H) } 
    = & \mathrm{arg min}_{\alpha > 0} \, \abs{ \frac{1}{ F_\alpha'(\sigma)} } \\
    = & \mathrm{arg max}_{\alpha > 0} \, \abs{ F_\alpha'(\sigma)}
\end{aligned}
\end{equation}


Using the relation~\eqref{eq:inverse_fct_rule}, we can show that the optimal hockey-stick divergence estimator is always near $\alpha=1$ which corresponds to the TV distance.

\begin{lem} \label{lem:max_sens}
For any $\sigma > 1$, as a function of $\alpha$, $\abs{F_\alpha'(\sigma)}$ has its maximum on the interval $[1,\ee^{\frac{1}{2 \sigma}}]$.
\end{lem}

\begin{figure}[h!]
\centering
\includegraphics[width=0.6\columnwidth]{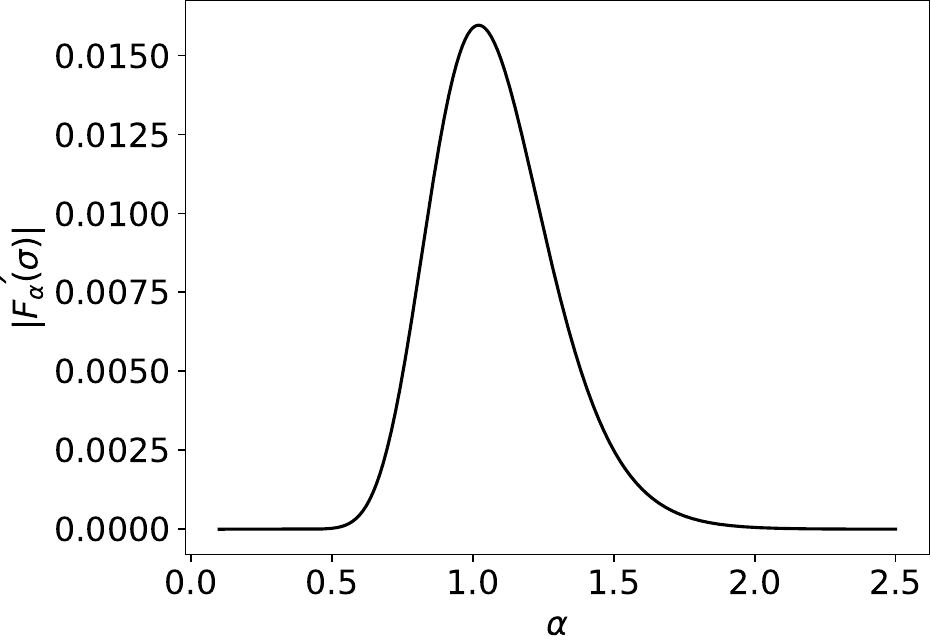}
\caption{The value of $\abs{ F_\alpha'(\sigma)}$ as a function of $\alpha$, when $\sigma=5$. We see that the optimal value is not far from $\alpha=1$, indicating that the choice $\alpha=1$ 
gives an estimate of $\sigma$ that is robust to errors. }
\label{fig:H_sensitivity}
\end{figure}

Figure~\ref{fig:H_sensitivity} illustrates numerically that the optimal value of $\alpha$ is not commonly far from 1.

\subsubsection*{Numerical Example}

Consider the two distributions $P \sim q \cdot \mathcal{N}(1,\sigma^2) + (1-q) \cdot \mathcal{N}(0,\sigma^2)$ and $Q \sim \mathcal{N}(0,\sigma^2)$ with the parameter values $q=1/4$ and $\sigma=0.3$. Estimating the $TV$-distance using $k=10^6$ samples from both $P$ and $Q$ and 20 bins we get the estimate $0.2256$. Using the fixed value  $q=1/4$, this translates to a  $\sigma$-estimate of $0.302$. Using Lemma~\ref{lem:canonne} we get for the TV distance a 99.99 \% - confidence interval  $[0.302 - 0.005, 0.302 + 0.005]$ which translates to a $\sigma$-interval $[0.285,0.32]$ so that 0.285 would be a 99.99 \%-confidence lower bound for $\sigma$.

\section{One-Shot Estimation Using Random Canaries} \label{sec:oneshot}

We next show results for one-shot estimation of DP guarantees using random canary gradients.
In the white-box setting, auditing of DP-SGD is often based on the assumption that the inserted canary gradient is (approximately) orthogonal against the rest of the per sample gradients~\cite{nasr2023tight}. The approach of~\cite{andrew2024one} leverages the fact that this is approximately obtained by sampling the gradients randomly since the inner products between random unit vectors diminish as the dimension increases. In~\cite{andrew2024one} it is shown that by taking the mean and variance of the inner products between the random canaries and model parameters, one can infer the $(\veps,\delta)$-DP guarantees and show that under mild assumptions the guarantees converge to the correct ones as the dimension $d \rightarrow \infty$. As we show, we can obtain a similar asymptotic result by applying Algorithm~\ref{alg:auditing_method} directly to the samples to estimate the $(\veps,\delta)$-guarantees instead of using means and variances and by assuming the Gaussian parametric form for the underlying DP noise.
To put our approach into perspective, we consider the same setting as in Thm.\;3.3 of~\cite{andrew2024one}.



\begin{thm} \label{thm:gaussian_data}
Denote the auditing training canaries $A_{\mathrm{train}} = \{x_1,\ldots, x_{n}\}$ and the auditing test canaries $A_{\mathrm{test}} = \{z_1,\ldots, z_{n}\}$, where $x_i$'s and $z_i$'s are i.i.d. uniformly sampled from the unit sphere $\mathbb{S}^{d-1}$. Let $n = \omega(1)$ (as a function of $d$) and $d = \omega(n^3 \log n)$.
Suppose $\mathcal{M}$ is such that for any dataset $D$ consisting of vectors in $\mathbb{R}^d$, $X \in \mathbb{R}^d$ denotes the sum of the vectors in $D$, and
$$
\mathcal{M}(D) = X + \sum_{x \in A_{\mathrm{train}}} x  +  Z, \quad Z \sim \mathcal{N}(0,\sigma^2 I_d).
$$
Let $\theta \sim \mathcal{M}(D)$ and $\norm{X}_2 = o\left( \sqrt{\frac{d}{n \log n}} \right)$. Denote the training and test scores by
$$
\wt{P} = \begin{bmatrix}
     \langle x_1, \theta  \rangle  \\
    \vdots \\
    \langle x_n, \theta  \rangle 
\end{bmatrix}, \quad
\wt{Q} = \begin{bmatrix}
     \langle z_1, \theta  \rangle  \\
    \vdots \\
    \langle z_n, \theta  \rangle 
\end{bmatrix}.
$$
Then, denoting $ \mathds{1}_n = \begin{bmatrix} 1 & \ldots & 1
\end{bmatrix}^T \in \mathbb{R}^n$, we have that
$$
\mathrm{TV}\left( \begin{bmatrix}
    \wt P \\ \wt Q
\end{bmatrix} , \mathcal{N} \left( \begin{bmatrix} \mathds{1}_n \\ 0 \end{bmatrix}, \sigma^2 I_{2n} \right) \right)
\rightarrow 0,
$$
as $d \rightarrow \infty$ in probability.
\end{thm}

Combining Theorem~\ref{thm:gaussian_data} with the convergence result of Theorem~\ref{thm:bin_convergence} we find that the $(\veps,\delta)$-distance between the histogram estimates of $\wt P$ and $\wt Q$ also converge to the DP guarantees of the Gaussian mechanism with noise scale $\sigma$.

\begin{cor} \label{cor:oneshot}
Denote by $\wh P$ and $\wh Q$ the histogram estimates obtained from the samples $\wt P$ and $\wt Q$, respectively, for some division of the real line.
Then, for all $\veps \in \mathbb{R}$, with an appropriate division of real line, we have that
$$
H_{\ee^{\veps}}( \wh P, \wh Q ) \rightarrow H_{\ee^{\veps}}\big( \mathcal{N}(1,\sigma^2 ), \mathcal{N}(1,\sigma^2 ) \big)
$$
as $d \rightarrow \infty$ in probability.
\end{cor}


\section{Experiments} \label{sec:experiments}

We illustrate the effectiveness of the histogram-based auditing in ML model auditing in both black-box and white-box setting, we consider a one hidden-layer feedforward network for MNIST classification~\cite{Lecun}, with hidder-layer width 200. 
We minimize the cross-entropy loss for all models, and all models are trained using the Adam optimizer~\cite{kingma2014adam} with the default initial learning rate 0.001. The clipping constant $C$ is set to 1.0.
To compute the theoretical $\veps$-upper bounds, we use the PRV accountant of the Opacus library~\cite{opacus2021}.

\subsection{Experiments on Black-Box Auditing} \label{sec:experiments_blackbox}

For the black-box auditing, we consider the method considered in~\cite{nasr2023tight} and depicted in Alg.~\ref{alg:blackbox_auditing}: using an auditing sample $(x',y')$, we draw $n$ samples of the loss function value evaluated on a model that is trained using DP-SGD on a dataset $D$ that does not include $z'$ and $n$ samples on dataset $D' = D  \cup (x',y')$. 
 
\begin{algorithm}[h!]
\caption{Black-box auditing method for DP-SGD.}
\begin{algorithmic}
\STATE{\textbf{Input:} Training dataset $D$, loss function $\ell$, canary input $(x',y')$, number of observations $T$.}
\STATE{Observations: $O \rightarrow [], O' \rightarrow []$.}
\STATE{Set: $D' = D \cup \{ (x',y') \}$.}
\FOR{ $t \in [n]:$ }
        \STATE{$\theta \rightarrow \mathcal{M}(D)$ (DP-SGD on the dataset $D$).}
        \STATE{$\theta' \rightarrow \mathcal{M}(D')$ (DP-SGD on the dataset $D'$).}
        \STATE{$O[t] \rightarrow \ell\big(\theta, (x',y') \big).$}
        \STATE{$O'[t] \rightarrow \ell\big(\theta', (x',y') \big).$}
\ENDFOR
\RETURN{$O,O'$.}
\end{algorithmic}
\label{alg:blackbox_auditing}
\end{algorithm}

We train the models with a random subset of 1000 samples from the training split of the MNIST dataset, and the additional sample $(x',y')$ is chosen randomly from the rest of the data.
We draw $n=10^4$ samples for both training and test scores. 
As a baseline method we consider the $\mu$-GDP auditing method cosidered in~\cite{nasr2023tight} that is obtained using a threshold inference and Clopper--Pearson confidence intervals for the FPR and FNR estimates. The threshold parameter is roughly optimized.
We also use the histogram-based estimation of the trade-off curves with Alg.~\ref{alg:auditing_method} and set the number of bins $k=10$ which, by Thm.~\ref{lem:canonne}, gives approximately $99 \%$ confidence intervals for the TV distance. 

\begin{figure}[h!]
\centering
\includegraphics[width=0.6\columnwidth]{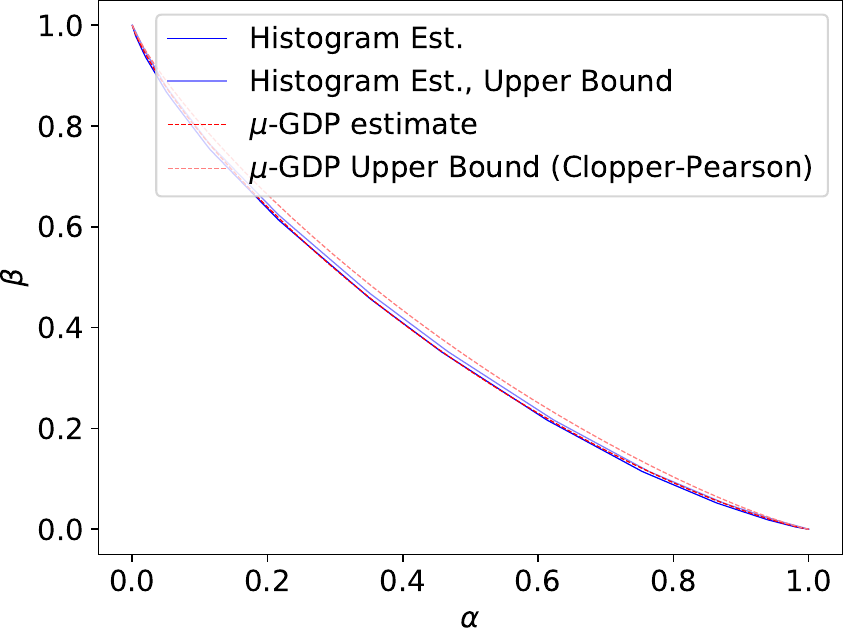} 
\caption{Estimated 99 \% - upper bound trade-off curves obtained using a) the thresholding and $\mu$-GDP and b) the histogram-based method using Alg.~\ref{alg:auditing_method}.
The empirical distributions of the losses are almost like Gaussians, which explains the fact that $\mu$-GDP auditing gives almost equally good estimates.
}
\label{fig:e3}
\end{figure}

\begin{figure}[h!]
\centering
\includegraphics[width=0.6\columnwidth]{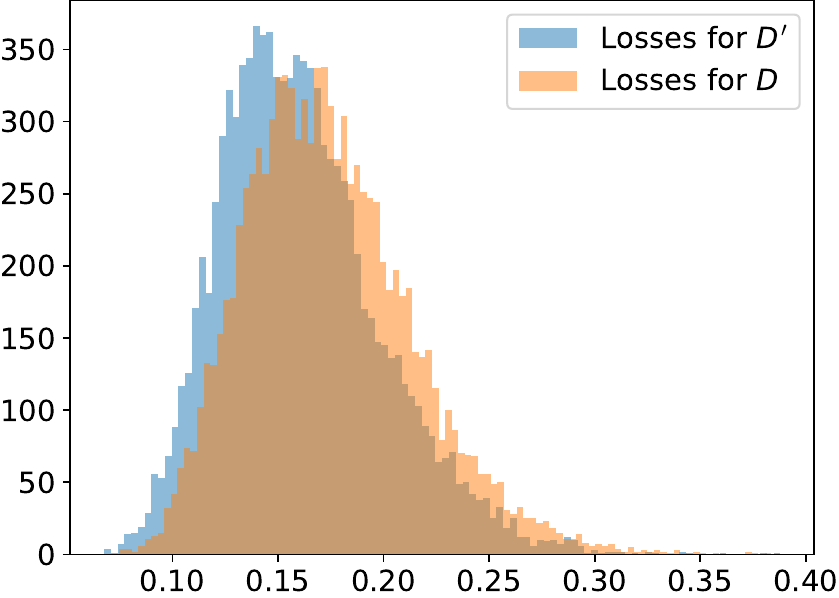} 
\caption{Histograms of the loss function values $\ell(\theta,x)$ at the end of the training,
when the model $\theta$ is trained using a) a dataset $D$ and b) dataset $D' = D \cup \{(x',y')\}$.
The empirical distribution deviate markedly from Gaussians.
}
\label{fig:e2}
\end{figure}

\begin{figure}[h!]
\centering
\includegraphics[width=0.6\columnwidth]{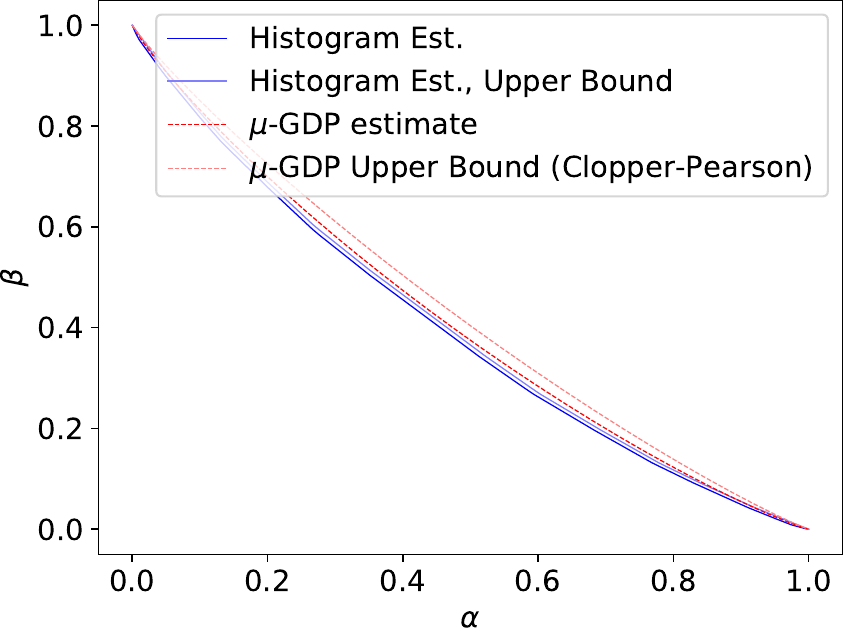} 
\caption{Estimated 99 \% - upper bound trade-off curves obtained using a) the thresholding and $\mu$-GDP and b) the histogram-based method using Alg.~\ref{alg:auditing_method}. The empirical distribution deviate from Gaussians (see above Fig.~\ref{fig:e2}), which explains the fact that $\mu$-GDP auditing gives worst estimates than the histogram-based approach.
}
\label{fig:e4}
\end{figure}

Figures~\ref{fig:e3} and~\ref{fig:e4} correspond to the trade-off curves obtained with two randomly chosen samples. 
For the first one, the histograms of the losses look approximately like Gaussians (the numerical skewness and kurtosis for the distributions are approximately 0.2 and 0.1, respectively) and we do not see a big difference in the trade-off curves (Fig.~\ref{fig:e3}) given by the two methods. However, for the other sample the histograms (see Fig.~\ref{fig:e2}) look less like Gaussians (the numerical skewness and kurtosis for the distributions are approximately 0.7 and 0.8, respectively) which explains that there is a bigger difference in the trade-off curves (Fig.~\ref{fig:e4}) given by the two methods.

\subsection{Experiments on White-Box Auditing} \label{sec:experiments_whitebox}

Lastly, we propose a heuristic white-box auditing method to estimate the privacy loss distributions of the underlying model training mechanism that can also be used to obtain estimates  for compositions, without any a priori information about the parameters of the training algorithm.

In~\cite{nasr2023tight} white-box auditing is carried out using Alg.~\ref{alg:whitebox_auditing} given in Appendix, such that the canary gradient is added with probability 1 at each iteration, i.e., the auditing neglects the effect of subsampling. Having an estimate of the $\mu$-GDP parameter gives then estimates of the so-called dominating pairs of distributions for the subsampled Gaussian mechanism in case the subsampling ratio $q$ is known. Using these, one can construct numerical privacy loss distributions (PLDs) and using FFT-based accounting methods~\cite{koskela2021tight,gopi2021} furthermore compute empirical $\delta(\veps)$-bounds also for compositions.

We consider the same setting of white-box auditing, however, we include the canaries  with the same probability as other gradients, and we carry out numerical estimation of the PLDs by estimating the distributions of inner product values in Alg.~\ref{alg:whitebox_auditing} using histograms, i.e., we calculate the discrete probabilities $\wh P$ and $\wh Q$ as in Alg.~\ref{alg:auditing_method}, and
then get the discrete-valued PLDs
$\omega_{\wh P/\wh Q}$ and $\omega_{\wh Q/\wh P}$, such that for $j \in [k]$, 
$$
\mathbb{P}\left( \omega_{\wh P/\wh Q} = \log \tfrac{\wh P_j}{\wh Q_j} \right) = \wh P_j
$$ 
and 
$$
\mathbb{P}\left( \omega_{\wh Q/\wh P} = \log \tfrac{\wh Q_j}{\wh P_j} \right) = \wh Q_j.
$$
We approximate the PLDs of a $c$-fold composition of the mechanism then by PLDs $\omega_{\wh P/\wh Q}^c$ and $\omega_{\wh Q/\wh P}^c$ that are given by $c$-fold self-convolutions of distributions $\omega_{\wh P/\wh Q}$ and $\omega_{\wh Q/\wh P}$, respectively, 
and obtain an 
 estimate $\wt \delta(\veps)$ of the privacy profile $\delta(\veps)$ of the $c$-fold composition of the mechanism as
\begin{equation} \label{eq:inte}
    \begin{aligned}
        \wt \delta(\veps)  &= \max \{ \mathbb{E}_{s \sim \omega_{\wh P/\wh Q}^c}[1 - \ee^{\veps-s}]_+, \\
& \quad  \quad  \quad  \quad         \mathbb{E}_{s \sim \wt \omega_{\wh Q/\wh P}^c}[1 - \ee^{\veps-s}]_+ \}.
    \end{aligned}
\end{equation}
The convolutions and the integrals~\eqref{eq:inte} are evaluated using the numerical method of~\cite{koskela2021tight}.

\begin{figure}[h!]
\centering
\includegraphics[width=0.6\columnwidth]{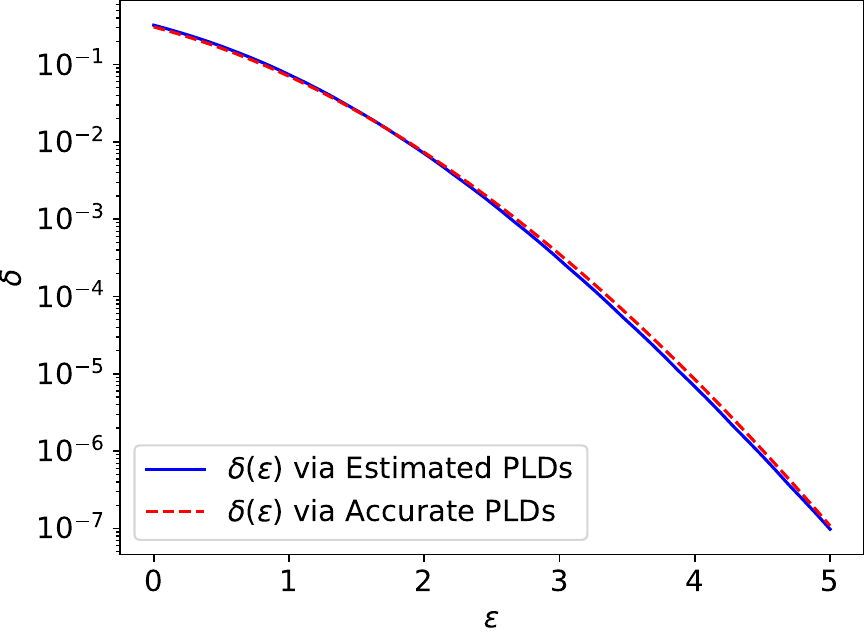} 
\caption{Comparison of the accurate privacy profile $\delta(\veps)$ and the estimated privacy profile that is computed using the discrete distributions $\wt P$ and $\wt Q$ obtained from the histogram estimates of $P$ and $Q$. Here $P \sim q \cdot \mathcal{N}(1,\sigma^2) + (1-q) \cdot \mathcal{N}(0,\sigma^2)$ and $Q \sim \cdot \mathcal{N}(0,\sigma^2)$, where $\sigma=2.0$ and $q=0.5$.
}
\label{fig:pld}
\end{figure}

Figure~\ref{fig:pld} shows results for a one-dimensional toy problem, where $P \sim q \cdot \mathcal{N}(1,\sigma^2) + (1-q) \cdot \mathcal{N}(0,\sigma^2)$ and $Q \sim \mathcal{N}(0,\sigma^2)$ with the parameter values $q=1/2$ and $\sigma=2.0$. 
We draw $n=10^5$ random samples from both $P$ and $Q$. We compute the $\delta(\veps)$-bounds for $c=10$ compositions and the accurate bounds are computed using the method of~\cite{gopi2021}.

\begin{figure}[h!]
\centering
\includegraphics[width=0.6\columnwidth]{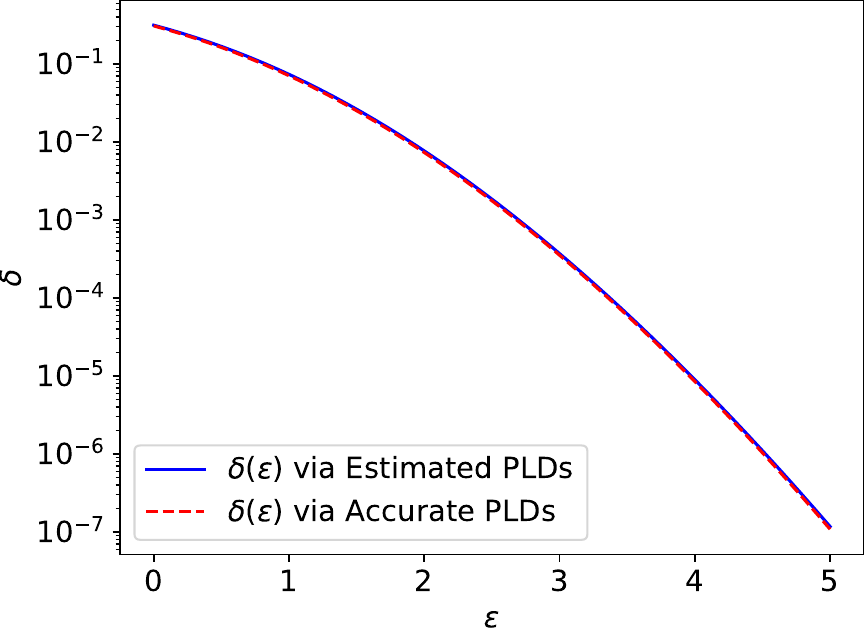}
\caption{ Comparison of the accurate privacy profile $\delta(\veps)$ and the estimated privacy profile that is computed using the discrete distributions $\wt P$ and $\wt Q$ obtained from the histogram estimates of $P$ and $Q$. Here samples from $P$ and $Q$ are obtained using inner products with random canaries (Alg.~\ref{alg:whitebox_auditing}).
}
\label{fig:e8}
\end{figure}

Similarly, Fig.~\ref{fig:e8} shows results for white-box auditing using Alg.~\ref{alg:whitebox_auditing} for the feedforward neural network, using a random subset of 1000 samples from the training split of the MNIST dataset. We use random normally distributed canaries and draw a new random canary vector at each step. We train $10^4$ models, each for 10 epochs, with a batch size of 500 and noise scale $\sigma=2.0$. We concatenate all the scores, giving in total $n=2 \cdot 10^5$ samples for both $P_S$ and $Q_S$ from which the histogram-estimates  $\wh P$ and $\wh Q$ are constructed.

\section{Conclusions} \label{sec:conclusions}

We have proposed a simple and practical technique to compute empirical estimates of DP privacy guarantees that does not require any a priori information about the underlying mechanism.
We have shown that our method can be seen as a generalization of the existing threshold membership inference auditing methods. 
One limitation of our method is that the reported $\veps$-estimates in the white-box setting are heuristic and we do not provide confidence intervals for them. To improve our methods, it would be important to find tighter confidence intervals for estimates of multinomial distributions (see, e.g.,~\cite{chafai2009confidence}). We leave this however for future work. To increase the computational efficiency, it will also be interesting to find conditions under which we can circumvent the assumption of the independence of the auditing score values when carrying out one-shot estimation and possibly give confidence intervals for $\veps$-lower bounds in that case. 

\clearpage

\bibliography{random_bib}

\newpage

\appendix

\section{Numerical Optimization to Find Accurate $\mu$-GDP parameter}

\begin{figure}[h!]
\centering
\includegraphics[width=0.6\columnwidth]{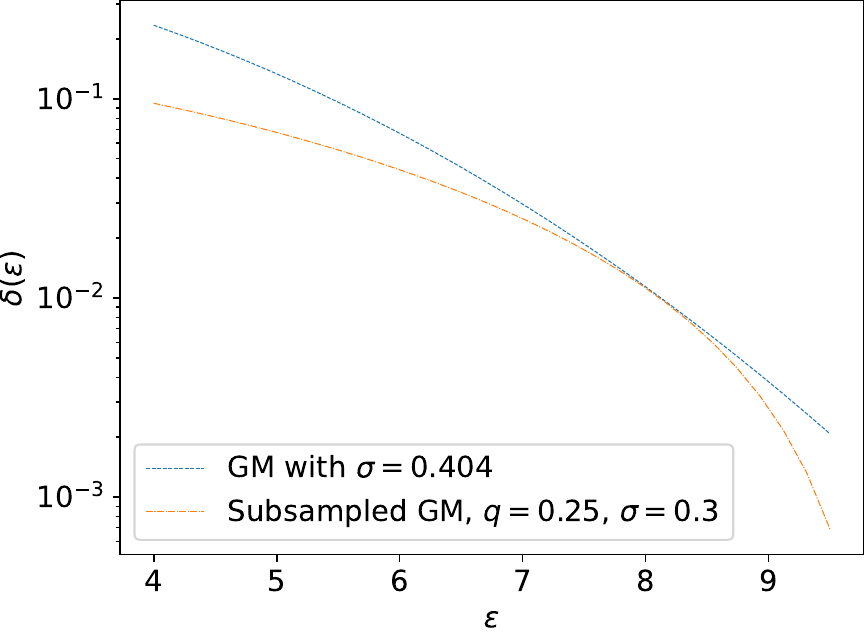} 
\caption{Adjusting the $\mu$-GDP parameter for the pair of distributions $P \sim q \cdot \mathcal{N}(1,\sigma^2) + (1-q) \cdot \mathcal{N}(0,\sigma^2)$ and $Q \sim \mathcal{N}(0,\sigma^2)$, where $\sigma=0.3$ and $q=0.25$. The tight $\mu$-GDP is given by $\mu=1/\sigma$, where $\sigma$ is the minimal value such that the  the privacy profile of the Gaussian mechanism with noise scale $\sigma$ is under the privacy profile $h(\alpha) = H_\alpha(P,Q)$. This value can be found, e.g., by solving the problem~\eqref{eq:problem}.
}
\label{fig:0}
\end{figure}

The numerical computation of the $\mu$-GDP parameter is be carried out such that
using the privacy profile
$$
\delta(\veps) = \max \{ H_{\ee^\veps}\big( P || Q \big), H_{\ee^\veps}\big( Q || P \big) \}
$$
where $P \sim q \cdot \mathcal{N}(1,\sigma^2) + (1-q) \cdot \mathcal{N}(0,\sigma^2)$ and $Q \sim \cdot \mathcal{N}(0,\sigma^2)$, 
we find a value of $\sigma$ for the Gaussian mechanism such that we search for a point, where the tangent and value of the privacy profiles $\delta(\veps)$ and $\delta_{\mathrm{Gauss},\mu}(\veps)$ are equal. To this end, we solve numerically the problem
\begin{equation} \label{eq:problem}
    \mathrm{arg min}_{\sigma} \min\nolimits_{\veps} \norm{\begin{bmatrix}
    \delta(\veps) \\ \tfrac{\dd }{\dd \veps}\delta(\veps)
\end{bmatrix}  - \begin{bmatrix}
    \delta_{\mathrm{Gauss},\sigma}(\veps) \\ \tfrac{\dd }{\dd \veps} \delta_{\mathrm{Gauss},\sigma}(\veps)
\end{bmatrix} }.
\end{equation}
Given a numerical solution $\wh \sigma$, the $\mu$-parameter is then given by $\mu = 1/\wh \sigma$.
Figure~\ref{fig:0} illustrates the result of this optimization.

\section{Proof of Lemma~\ref{lem:TV_to_eps_delta_bound}} 

\begin{lem} 
Denote $P,Q$ probability distributions on the same probability space. Suppose 
$$
\mathrm{TV}(P,\wt P) \leq \tau
$$
and
$$
\mathrm{TV}(Q,\wt Q) \leq \tau
$$
for some $\tau \geq 0$. Then, for all $\veps \in \mathbb{R}$,
$$
H_{\ee^{\veps}}(P,Q) \leq H_{\ee^{\veps}}(\wt P, \wt Q) + (1+\ee^{\veps}) \cdot \tau.
$$
\begin{proof}
Using the inequality $\ \leq [a]_+ + [b]_+$ that holds for all $a,b \in \mathbb{R}$, we have that
\begin{equation*}
    \begin{aligned}
      H_{\ee^{\veps}}(P,Q) & = \int [P(t) - \ee^{\veps} Q(t)]_+ \, \dd t \\
      &=  \int  [ P(t) - \wt P(t) + \ee^{\veps}(\wt Q(t) - Q(t))  + \wt P(t) - \ee^{\veps} \wt Q(t)]_+ \, \dd t \\
&\leq   \int \ [ P(t) - \wt P(t)]_+ \, \dd t  + \ee^{\veps} \int [\wt Q(t) - Q(t)]_+ \, \dd t
+ \int [\wt P(t) - \ee^{\veps} \wt Q(t)]_+ \, \dd t \\
&=  TV(P,\wt P) + \ee^{\veps} TV(\wt Q,Q) + H_{\ee^{\veps}}(\wt P,\wt Q) \\
&\leq (1+\ee^{\veps}) \cdot \tau  + H_{\ee^{\veps}}(\wt P,\wt Q).
\end{aligned}
\end{equation*}

\end{proof}
\end{lem}

\section{Trade-Off Function for the Laplace Mechanism}

The accurate trade-off function of the Laplace mechanism is given in Lemma A.6 of~\cite{dong2022gaussian}
\begin{equation*}
    \begin{aligned}
        & T \big( \mathrm{Lap}(0,1),\mathrm{Lap}(\mu,1) \big)(\alpha) = \\ & \begin{cases}
    1-\ee^{\mu} \alpha, \quad & \alpha < \ee^{-\mu}/2. \\
    \ee^{-\mu}/4\alpha, \quad & \ee^{-\mu}/2 \leq \alpha \leq 1/2, \\
        \ee^{-\mu}(1-\alpha)    , \quad & \alpha \geq 1/2,
\end{cases}
    \end{aligned}
\end{equation*}

\section{Illustration: Conversion Between the Noise Parameter and TV Distance for the Gaussian Mechanism}

\begin{figure}[h!]
\centering
\includegraphics[width=0.6\columnwidth]{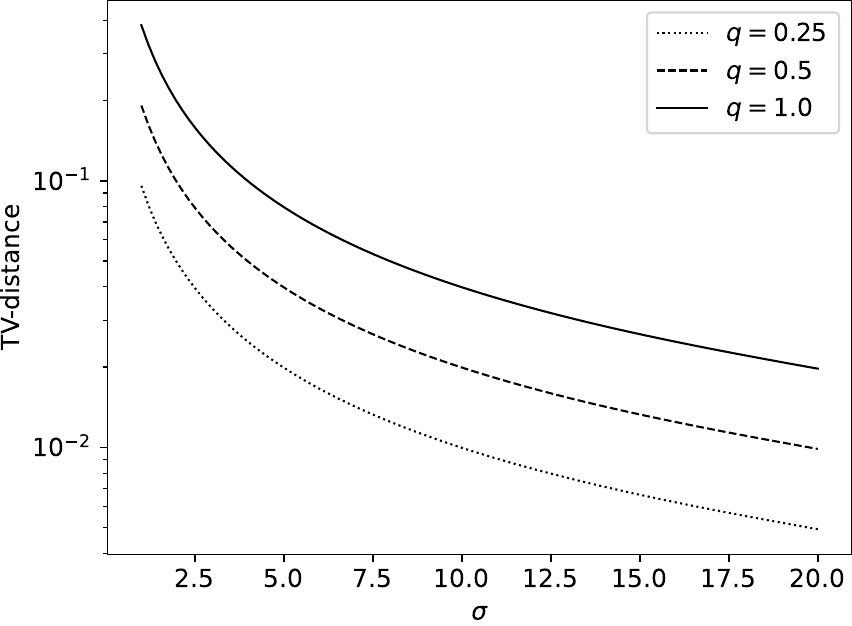}
\caption{Relationship between TV distance $TV(P,Q)$ and $\veps$ when $\delta=10^{-5}$ for the Gaussian mechanism.}
\label{fig:2}
\end{figure}

Figure~\ref{fig:2} shows the TV distance $TV(P,Q)$ when $P \sim q \cdot \mathcal{N}(1,\sigma^2) + (1-q) \cdot \mathcal{N}(0,\sigma^2)$ and $Q \sim \mathcal{N}(0,\sigma^2)$ for three different values of $q$ and for varying values of $\sigma$.

Using the conversion from the TV distance to $\sigma$, we can also convert the confidence intervals for the confidence interval of $TV(P,Q)$ in case we know the subsampling parameter $q$.

\section{Proof of Lemma~\ref{lem:kairouz}} \label{Asec_lem:kairouz}

\begin{lem} 
Consider two probability distributions $P$ and $Q$ and distributions $P$ and $Q_2$ obtained with two-bin frequency histograms defined by a threshold $\tau \in \mathbb{R}$. Suppose the underlying mechanism $\mathcal{M}$ is $(\veps,\delta)$-DP for some $\veps \geq 0$ and $\delta \in [0,1]$. Then, asymptotically,
\begin{equation} \label{eq:ineqineq}
\max\{ H_{\ee^{\veps}}\big( P_2 || Q_2 \big), H_{\ee^{\veps}}\big( Q_2 || P_2 \big) \} \leq \delta,
\end{equation}
where
\begin{equation*}
        \veps = \max \left\{ \log \frac{\mathrm{TPR} - \delta}{\mathrm{FPR}}, \log \frac{\mathrm{TNR} - \delta}{\mathrm{FNR}} \right\}.
\end{equation*}
and $\mathrm{TPR}$ and $\mathrm{FPR}$ are as defined in Eq.~\eqref{eq:TPRFPR} and $\mathrm{FNR} = 1 - \mathrm{TPR}$ and $\mathrm{TNR} = 1 - \mathrm{FPR}$.
\begin{proof}
We have that $\wh  P$ and $\wh  Q$ are now discrete distributions with binary values, such that
$\wh P=(p_1,p_2)$, where $p_1 = \mathrm{TPR}$, $p_2=\mathrm{FNR}$ and 
$\wh Q=(q_1,q_2)$, where $q_1 = \mathrm{FPR}$, $q_2=\mathrm{TNR}$. Assuming $p_1 \geq q_1$, i.e., $q_2 \geq p_2$, we have that
\begin{equation} \label{eq:first_H}
H_{\ee^{\veps}}\big(\wh  P ||\wh  Q \big)
= [p_1 - \ee^{\veps} q_1]_+ + [p_2 - \ee^{\veps} q_2]_+ = p_1 - \ee^{\veps} q_1
\end{equation}
and
\begin{equation} \label{eq:second_H}
H_{\ee^{\veps}}\big(\wh  Q ||\wh  P \big)
= [q_1 - \ee^{\veps} p_1]_+ + [q_2 - \ee^{\veps} p_2]_+ = q_2 - \ee^{\veps} p_2.
\end{equation}
Setting Eq.~\eqref{eq:first_H} and~\eqref{eq:second_H} equal to $\delta$ gives $\veps = \frac{p_1 - \delta}{q_1}$
and $\veps = \frac{q_2 - \delta}{p_2}$, respectively.
Taking maximum of $H_{\ee^{\veps}}\big(\wh  P || \wh Q \big)$ and $H_{\ee^{\veps}}\big(\wh  Q || \wh  P \big)$ for $\delta$
is equivalent to taking the maximum of $\frac{p_1 - \delta}{q_1}$ and $\frac{p_1 - \delta}{q_1}$ for $\veps$. The inequality in~\eqref{eq:ineqineq} follows from the data-processing inequality (same post-processing applied to $P$ and $Q$).
\end{proof}
\end{lem}

\section{Proof of Theorem~\ref{thm:bin_convergence}} \label{Asec_thm:bin_convergence}

\begin{thm} 
Let $P$ and $Q$ be one-dimensional probability distributions with differentiable density functions $P(x)$ and $Q(x)$, respectively, and
consider the histogram-based density estimation described above. Draw $n$ samples both from $P$ and $Q$, giving density estimators $\wh P = (\wh P_1, \ldots, \wh P_k)$ and $\wh Q = (\wh Q_1, \ldots, \wh Q_k)$, respectively.
Let the bin width be chosen as
\begin{equation*} 
h_n = \left( \frac{12}{\int P'(x)^2 \, \dd x + \int Q'(x)^2 \, \dd x} \right)^{\frac{1}{3}} n^{-\frac{1}{3}}
\end{equation*}
Then, for any $\alpha \geq 0$, the numerical hockey-stick divergence $H_\alpha(\wh P || \wh Q)$ convergences in expectation to 
$H_\alpha(P || Q)$ with rate $\mathcal{O}(n^{-1/3})$, i.e.,
$$
\mathbb{E}\abs{H_\alpha(\wh P || \wh Q) - H_\alpha(P || Q) } = \mathcal{O}(n^{-1/3}),
$$
where the expectation is taken over the random draws for constructing $\wh P$ and $\wh Q$.
\begin{proof}
Define the piece-wise continuous functions $\wh P(x)$ and $\wh Q(x)$ such that $\wh P(x) =  \wh P_\ell/h$, if $x \in \textrm{Bin}_\ell$ and similarly for $\wh Q(x)$. To analyse the error in the hockey-stick divergence estimate we can use $\wh P(x)$ and $\wh Q(x)$ since
\begin{equation*} 
	\begin{aligned}
H_\alpha( \wh P || \wh Q ) & = \sum_{\ell=1}^N [\wh P_\ell - \alpha \cdot \wh Q_\ell]_+ \\
& = \int [\wh P(x) - \alpha \cdot \wh Q(x)]_+ \, \dd x \\
& =H_\alpha( \wh P(x) || \wh Q(x) ) .
\end{aligned}
\end{equation*}
We can bound the divergence $H_\alpha(\wh P || \wh Q)$ as follows:
\begin{equation} \label{eq:ineq_Hpkqk}
	\begin{aligned}
 H_\alpha(\wh P &|| \wh Q) \\ 
= &  \int  [ \wh P(x) - \alpha \cdot \wh Q(x)]_+ \, \dd x \\
= & \int  [ \wh P(x) - P(x) - \alpha \cdot (\wh Q(x)-Q(x)) + P(x) - \alpha \cdot Q(x)]_+ \, \dd x \\
\leq & \int  \abs{ \wh P(x) - P(x) } \, \dd x + \alpha  \int   \abs{ \wh Q(x) - Q(x) } \, \dd x  + \int  [  P(x) - \alpha \cdot Q(x)]_+ \, \dd x \\
\leq & \sqrt{ \int  (\wh P(x) - P(x))^2 \, \dd x} + \alpha \sqrt{ \int  (\wh Q(x) - Q(x))^2 \, \dd x}  + \int  [  P(x) - \alpha \cdot Q(x)]_+ \, \dd x \\
= & \sqrt{ \int  (\wh P(x) - P(x))^2 \, \dd x}  + \alpha \sqrt{ \int  (\wh Q(x) - Q(x))^2 \, \dd x}
+  H_\alpha(P  || Q ) \\
\leq & \max \{1,\alpha \} \cdot \bigg( \sqrt{ \int  (\wh P(x) - P(x))^2 \, \dd x}  + \sqrt{ \int  (\wh Q(x) - Q(x))^2 \, \dd x} \bigg)
+  H_\alpha(P  || Q ),
	\end{aligned}
\end{equation}
where the first inequality follows from the fact that $[ a + b]_+ \leq \abs{a} + [ b]_+ $ for all $a,b \in \mathbb{R}$,
and the second inequality follows from the H\"older inequality.

Similarly, carrying out the same calculation starting from $H_\alpha(P || Q)$, we have
\begin{equation} \label{eq:ineq_Hpkqk2}
	\begin{aligned}
H_\alpha(P  || Q) 
& = \int [ P(x) - \alpha \cdot Q(x)]_+ \, \dd x \\
&\leq  \int  \abs{ \wh P(x) - P(x) } \, \dd x + \alpha  \int  \abs{ \wh Q(x) - Q(x) } \, \dd x  + \int  [  \wh P(x) - \alpha \cdot \wh Q(x)]_+ \, \dd x \\
& =  \int  \abs{ \wh P(x) - P(x) } \, \dd x  + \alpha  \int   \abs{ \wh Q(x) - Q(x) } \, \dd x  +  H_\alpha(\wh P || \wh Q) \\
&\leq  \max \{1,\alpha \} \cdot \bigg( \sqrt{ \int  (\wh P(x) - P(x))^2 \, \dd x}  + \sqrt{ \int  (\wh Q(x) - Q(x))^2 \, \dd x} \bigg) +  H_\alpha(\wh P || \wh Q )
\end{aligned}
\end{equation}
From the inequalities~\eqref{eq:ineq_Hpkqk} and~\eqref{eq:ineq_Hpkqk2} it follows that
\begin{equation} \label{eq:ineq_Hpkqk3}
\begin{aligned}
 \abs{H_\alpha(\wh P || \wh Q) - H_\alpha(P || Q) }  \leq \max \{1,\alpha \} \cdot  \bigg( \bigg( \int  (\wh P(x) - P(x))^2 \, \dd x \bigg)^{\frac{1}{2}}  + \bigg( \int  (\wh Q(x) - Q(x))^2 \, \dd x \bigg)^{\frac{1}{2}} \bigg).
\end{aligned}
\end{equation}
Taking the expectation over the random draws from $P$ and $Q$ and applying Jensen's inequality to the the square root function, we get
\begin{equation} \label{eq:ineq_Hpkqk4}
\begin{aligned}
 \mathbb{E} \abs{H_\alpha(\wh P || \wh Q) - H_\alpha(P || Q) }  &\leq \max \{1,\alpha \} \cdot \bigg( \bigg( \int \mathbb{E} (\wh P(x) - P(x))^2 \, \dd x \bigg)^{\frac{1}{2}}  + \bigg( \int \mathbb{E} (\wh Q(x) - Q(x))^2 \, \dd x\bigg)^{\frac{1}{2}} \bigg) \\
& \leq \sqrt{2} \max \{1,\alpha \} \bigg(  \int \mathbb{E} (\wh P(x) - P(x))^2 \, \dd x + \int \mathbb{E} (\wh Q(x) - Q(x))^2 \, \dd x \bigg)^{\frac{1}{2}},
\end{aligned}
\end{equation}
where the second inequality follows from the inequality $\sqrt{a} + \sqrt{b} \leq \sqrt{2} \sqrt{a+b}$ which holds for any $a,b \geq 0$.
From the derivations of Sec.\;3 of~\cite{scott1979optimal} we have that
\begin{equation} \label{eq:intermediate_scott}
	\begin{aligned}
 \int \mathbb{E}  & (\wh P(x) - P(x))^2 \, \dd x + \int \mathbb{E} (\wh Q(x) - Q(x))^2 \, \dd x \\
 = & \frac{2}{n \cdot h} + \frac{1}{12} h^2 \left[ \int P'(x)^2 \, \dd x + \int Q'(x)^2 \, \dd x \right]  + \mathcal{O}\left(\frac{1}{n} + h^3 \right)
\end{aligned}
\end{equation}
Minimizing the first two terms on the right-hand side of~\eqref{eq:intermediate_scott} with respect to $h$ gives the expression of $h_n$ and furthermore, with this choice $h_n$, we have that 
$$
 \int \mathbb{E} (\wh P(x) - P(x))^2 \, \dd x + \int \mathbb{E} (\wh Q(x) - Q(x))^2 \, \dd x 
 = \mathcal{O}\left(n^{- \frac{2}{3}} \right).
$$
which together with the inequality~\eqref{eq:ineq_Hpkqk4} shows that 
$$
\mathbb{E} \abs{H_\alpha(\wh P || \wh Q) - H_\alpha(P || Q) } = \mathcal{O}(n^{-\frac{1}{3}}).
$$
\end{proof}
\end{thm}

\section{Proof of Theorem~\ref{thm:gaussian_data}} \label{Asec_lem:gaussian_data}

We first state some auxiliary results needed for the proof. Recall first the following result by~\cite{cai2013distributions} which states that maximal angle between $n$ random unit vectors goes to $\frac{\pi}{2}$ in probability as the dimension $d$ grows, in case $\frac{ \log n}{d} \rightarrow 0$ (see Thm.\;5 in~\cite{cai2013distributions}).

\begin{lem} \label{lem:inner_p3}
Let $x_1, \ldots, x_n$ be independently uniformly chosen random vectors from the unit sphere $\mathbb{S}^{d-1}$. Let $d = d_n \rightarrow \infty$ satisfy $\frac{ \log n}{d} \rightarrow 0$ as $n \rightarrow \infty$. Denote $\theta_{ij}$ the angle between the vectors $x_i$ and $x_j$. Then,
$$
\max_{1 \leq i < j \leq n} \abs{ \theta_{ij} - \frac{\pi}{2}} \rightarrow 0
$$
in probability as $n \rightarrow \infty$.
\end{lem}

Looking at the proof of Thm.\;5 of~\cite{cai2013distributions}, we obtain the following convergence speed for $\max_{1 \leq i < j \leq n} \abs{ \theta_{ij} - \frac{\pi}{2}}$.

\begin{lem} \label{lem:inner_p4}
Let the assumptions of Lemma~\ref{lem:inner_p3} hold. Then,
$$
\sqrt{\frac{d}{\log n}} \cdot \max_{1 \leq i < j \leq n} \abs{ \theta_{ij} - \frac{\pi}{2}} \rightarrow 4.
$$
\begin{proof}
This result corresponds to Corollary 2.1 of~\cite{cai2013distributions}. It can be shown similarly as in Thm.\;5 of~\cite{cai2013distributions}, i.e., by replacing in the proof of Thm.\;1 of~\cite{cai2012phase} $L_n$ and $\abs{\rho_{ij}}$ by $\max_{1 \leq i < j \leq n} \abs{ \theta_{ij} - \frac{\pi}{2}}$ and $\rho_{ij}$, respectively.
\end{proof}
\end{lem}
The convergence of the cosine angles trivially follows from the Lipschitz continuity of the cosine function.
\begin{cor} \label{cor:cai}
    Let $x_1, \ldots, x_n$ be independently uniformly chosen random vectors from the unit sphere $\mathbb{S}^{d-1}$. Let $d = d_n \rightarrow \infty$ satisfy $\frac{ \log n}{d} \rightarrow 0$ as $n \rightarrow \infty$. Denote $\rho_{ij}$ the cosine angle between the vectors $x_i$ and $x_j$. Then,
$$
\max_{1 \leq 1 < j \leq n} \abs{ \rho_{ij} } \rightarrow 0
$$
in probability as $n \rightarrow \infty$.
\begin{proof}
The results follows from Lemma~\ref{lem:inner_p3} and from the fact that cosine function is 1-Lipschitz:
$$
\abs{ \rho_{ij} } = 
\abs{ \langle x_i, x_j \rangle } = 
\abs{ \cos \theta_{ij}  }
=
\abs{ \cos \theta_{ij} - \cos \frac{\pi}{2} }
\leq \abs{ \theta_{ij} -\frac{\pi}{2} }.
$$
\end{proof}
\end{cor}

We will also need the following result by~\cite{devroye2018total} for the TV distance between two Gaussians with equal means (see Thm.\;1.1 in~\cite{devroye2018total}).

\begin{lem} \label{lem:devroye}
Let $\mu \in \mathbb{R}^d$, $\Sigma_1$ and $\Sigma_2$ be positive definite $d \times d$ matrices, and $\lambda_1, \ldots, \lambda_d$ denote the eigenvalues of $\Sigma_2^{-1} \Sigma_1 - I$. Then,
$$
\mathrm{TV}\big( \mathcal{N}(\mu,\Sigma_1), \mathcal{N}(\mu,\Sigma_2) \big) \leq \frac{3}{2} \min \left\{1, \sqrt{\sum\nolimits_{i=1}^d \lambda_i^2} \right\}.
$$
\end{lem}

\begin{thm}
Denote the auditing training canaries $A_{\mathrm{train}} = \{x_1,\ldots, x_{n}\}$ and the auditing test canaries $A_{\mathrm{test}} = \{z_1,\ldots, z_{n}\}$, where $x_i$'s and $z_i$'s are i.i.d. uniformly sampled from the unit sphere $\mathbb{S}^{d-1}$. Let $n = \omega(1)$ (as a function of $d$) and $d = \omega(n^3 \log n)$.
Suppose $\mathcal{M}$ is such that for any dataset $D$ consisting of vectors in $\mathbb{R}^d$, $X \in \mathbb{R}^d$ denotes the sum of the vectors in $D$, and
$$
\mathcal{M}(D) = X + \sum_{x \in A_{\mathrm{train}}} x  +  Z, \quad Z \sim \mathcal{N}(0,\sigma^2 I_d).
$$
Let $\theta \sim \mathcal{M}(D)$ and $\norm{X}_2 = o\left( \sqrt{\frac{d}{n \log n}} \right)$. Denote the training and test scores by
$$
\wt{P} = \begin{bmatrix}
     \langle x_1, \theta  \rangle  \\
    \vdots \\
    \langle x_n, \theta  \rangle 
\end{bmatrix}, \quad
\wt{Q} = \begin{bmatrix}
     \langle z_1, \theta  \rangle  \\
    \vdots \\
    \langle z_n, \theta  \rangle 
\end{bmatrix}.
$$
Then, denoting $ \mathds{1}_n = \begin{bmatrix} 1 & \ldots & 1
\end{bmatrix}^T \in \mathbb{R}^n$, we have that
$$
\mathrm{TV}\left( \begin{bmatrix}
    \wt P \\ \wt Q
\end{bmatrix} , \mathcal{N} \left( \begin{bmatrix} \mathds{1}_n \\ 0 \end{bmatrix}, \sigma^2 I_{2n} \right) \right)
\rightarrow 0,
$$
as $d \rightarrow \infty$ in probability.
\begin{proof}
Denote $\theta = X + \sum_{x \in A_{\mathrm{train}}} x  +  Z$, where $Z \sim \mathcal{N}(0,\sigma^2 I_d)$.
We see that for any $x_i \in A_{\mathrm{train}}$,
\begin{equation} \label{eq:train_scores}
\begin{aligned}
    S\big(x_i, \theta \big) &= x_i^T \left( X + \sum\limits_{x \in A_{\mathrm{train}}} x  + Z \right) \\
&= x_i^T X + 1 + \sum\limits_{x \in A_{\mathrm{train}}, \,x \neq x_i} x_i^T x +  x_i^T Z,
\end{aligned}
\end{equation}
and for any $z_i \in A_{\mathrm{test}}$,
\begin{equation} \label{eq:test_scores}
\begin{aligned}
S\big(z_i, \theta \big) &= z_i^T \left( X + \sum\limits_{x \in A_{\mathrm{train}}} x  + Z \right) \\
&= z_i^T X + \sum\limits_{x \in A_{\mathrm{train}}} z_i^T x + z_i^T Z.
\end{aligned}
\end{equation}
From Eq.~\eqref{eq:train_scores} an~\ref{eq:test_scores} we see that 
$$
\begin{bmatrix}
    \wt P \\ \wt Q
\end{bmatrix} = C^T X + \begin{bmatrix} \mathds{1}_n \\ 0 \end{bmatrix} + \tau + C^T Z,
$$
where 
$$
C = \begin{bmatrix}
x_1 & \ldots & x_n & z_1 & \ldots & z_n 
\end{bmatrix} 
$$
and
$$
\tau_i = \begin{cases}
    \sum\limits_{x \in A_{\mathrm{train}}, \,x \neq x_i} x_i^T x, & \quad   1 \leq i \leq n \\
    \sum\limits_{x \in A_{\mathrm{train}}} z_i^T x, & \quad  n < i \leq 2n
\end{cases}
$$

Denote the maximum absolute cosine angle between the vectors $\tfrac{X}{\norm{X}}, x_1, \ldots, x_n, z_1, \ldots, z_n$ by $\rho_{\max}$. We easily see that $\rho_{\max}$ has the same distribution as the maximum absolute cosine angle between $2n +1$ vectors uniformly sampled from the unit sphere $\mathbb{S}^{d-1}$.
Also, 
we have that for all $x_i \in A_{\mathrm{train}}$,
$$
\abs{\sum\nolimits_{x \in A_{\mathrm{train}}, \,x \neq x_i} x_i^T x} \leq n \cdot \rho_{\max}
$$
and for all $z_i \in A_{\mathrm{test}}$,
$$
\abs{\sum\nolimits_{x \in A_{\mathrm{train}}} \wh z_i^T x } \leq  n \cdot \rho_{\max}.
$$
Thus,
\begin{equation} \label{eq:z2}
\norm{\tau}_2 \leq \sqrt{2} n^{3/2} \rho_{\max}.
\end{equation}
Moreover, we have that 
\begin{equation} \label{eq:z3}
\norm{C^T X}_2 \leq \norm{X}_2 \sqrt{2n} \cdot \rho_{\max}.
\end{equation}
Moreover, by Lemma~\ref{lem:inner_p4}, we have that 
\begin{equation} \label{eq:z4}
\sqrt{\frac{d}{\log 2n+1}} \cdot \rho_{\max} \rightarrow 4
\end{equation}
as $d \rightarrow \infty$ in probability (since $n = \omega(1)$ as a function of $d$).
Combining Eq.~\eqref{eq:z4} with the bounds~\eqref{eq:z2} and~\eqref{eq:z3}, we see that
$\norm{\tau}_2 \rightarrow 0$ as $d \rightarrow \infty$ in probability in case 
$d = \omega(n^3 \log n)$ and $\norm{C^T X}_2 \rightarrow 0$ as $d \rightarrow \infty$ in probability in case $\norm{X}_2 = o\left( \sqrt{\frac{d}{n \log n}} \right)$.

We then bound using the triangle inequality as
\begin{equation} \label{eq:TV_b}
    \begin{aligned}
        & \mathrm{TV}\left( \begin{bmatrix}
    \wt P \\ \wt Q
\end{bmatrix} , \mathcal{N} \left( \begin{bmatrix} \mathds{1}_n \\ 0 \end{bmatrix}, \sigma^2 I_{2n} \right) \right) \\
= &  \mathrm{TV}\left( C^T X + \begin{bmatrix} \mathds{1}_n \\ 0 \end{bmatrix} + \tau + C^T Z , \mathcal{N} \left( \begin{bmatrix} \mathds{1}_n \\ 0 \end{bmatrix}, \sigma^2 I_{2n} \right) \right) \\
\leq & \mathrm{TV}\left( \begin{bmatrix} \mathds{1}_n \\ 0 \end{bmatrix} + C^TZ , \mathcal{N} \left( \begin{bmatrix} \mathds{1}_n \\ 0 \end{bmatrix}, \sigma^2 I_{2n} \right) \right)  + \mathrm{TV}\left(   C^T X + \tau + \begin{bmatrix} \mathds{1}_n \\ 0 \end{bmatrix} + C^TZ, \begin{bmatrix} \mathds{1}_n \\ 0 \end{bmatrix} + C^TZ \right)
    \end{aligned}
\end{equation}
We next use Lemma~\ref{lem:devroye} to show the convergence of the first term on the right hand side of the inequality~\eqref{eq:TV_b}.
Clearly, since $Z \sim \mathcal{N}(0,\sigma^2 I_d)$, we have that $C^TZ \sim \mathcal{N}(0, \sigma^2 C^T C)$. 
We next use Lemma~\ref{lem:devroye} with $\mu = \begin{bsmallmatrix} \mathds{1}_n \\ 0 \end{bsmallmatrix}$, $\Sigma_1 = C^TC$ and $\Sigma_2 = \sigma^2 I_{2n}$.
Denoting $\lambda_1, \ldots, \lambda_{2n}$ the eigenvalues of the matrix $\Sigma_2^{-1} \Sigma_1 = (\sigma^2 I)^{-1} \sigma^2 C^TC - I = C^TC - I$, we have that 
\begin{equation*} 
    \begin{aligned}
 \sum_{i=1}^{2n} \lambda_i^2 = &\norm{C^TC - I}_F^2 \\
=  & \sum_{ a,b \in \{x_1, \ldots, x_n, z_1, \ldots, z_n\}, a \neq b } (a^T b)^2  \\
\leq & (2n)^2 \rho_{\max}^2.
 \end{aligned}
\end{equation*}
By the assumption $d = \omega(n^3 \log n)$ and Lemma~\ref{lem:inner_p4} and Eq.~\eqref{eq:z4} we have that 
$\sum_{i=1}^{2n} \lambda_i^2 \rightarrow 0$ as $d \rightarrow \infty$ in probability, and therefore by Lemma~\ref{lem:devroye},
$$
\mathrm{TV}\left( \begin{bmatrix} \mathds{1}_n \\ 0 \end{bmatrix} + C^TZ , \mathcal{N} \left( \begin{bmatrix} \mathds{1}_n \\ 0 \end{bmatrix}, \sigma^2 I_{2n} \right) \right) \rightarrow 0
$$
as $d \rightarrow \infty$ in probability.

To show the convergence of the second term on the right hand side of the inequality~\eqref{eq:TV_b}, we again use the fact that $\norm{C^TC - I}_F^2 \rightarrow 0$ as $d \rightarrow \infty$ in probability, the unitary invariance of the total variation distance and the fact that $\norm{\tau}_2 \rightarrow 0$ and $\norm{C^T X}_2 \rightarrow 0$ as $d \rightarrow \infty$ in probability. 
\end{proof}

\end{thm}

\section{Proof of Corollary~\ref{cor:oneshot}}

\begin{cor} 
Suppose the assumptions of Theorem~\ref{thm:gaussian_data} hold.
Denote by $\wh P$ and $\wh Q$ the histogram estimates obtained from the samples $\wt P$ and $\wt Q$, respectively, for some division of the real line.
Then, for all $\veps \in \mathbb{R}$, with an appropriate division of real line, we have that
$$
H_{\ee^{\veps}}( \wh P, \wh Q ) \rightarrow H_{\ee^{\veps}}\big( \mathcal{N}(1,\sigma^2 ), \mathcal{N}(1,\sigma^2 ) \big)
$$
as $d \rightarrow \infty$ in probability.
\begin{proof}
Consider an equidistant division of the real line into intervals, with some bin width $h$, and suppose the probability estimates $\wh P$ and $\wh Q$ are obtained from the histogram estimates of the samples $\wt P$ and $\wt Q$, respectively. Denote by $N_1$ and $N_0$ the histogram estimates from using $n$ samples from $\mathcal{N}(0,\sigma^2)$ and  $\mathcal{N}(1,\sigma^2)$, respectively.
Similarly to the proof of Lemma~\ref{lem:TV_to_eps_delta_bound}, we have that
\begin{equation} \label{eq:ineq_z}
    \begin{aligned}
        H_{\ee^{\veps}}( \wh P, \wh Q ) & \leq H_{\ee^{\veps}}( N_0 , N_1 ) \\
        & + (1+\ee^{\veps}) \big( \mathrm{TV}(\wh P,N_1) + \mathrm{TV}(\wh Q,N_0) \big).
    \end{aligned}
\end{equation}
We obtain the sequences of masses $(\wh P,\wh Q)$ and $(N_0,N_1)$ by applying the same post-processing to the vectors 
$\begin{bsmallmatrix} \wt P \\ \wt Q \end{bsmallmatrix}$ and 
$\mathcal{N} \left( \begin{bsmallmatrix} \mathds{1}_n \\ 0 \end{bsmallmatrix}, \sigma^2 I_{2n} \right) $, 
respectively. 

Therefore
$$
\mathrm{TV}(\wh P,N_1) + \mathrm{TV}(\wh Q,N_0) \leq \mathrm{TV}\left( \begin{bmatrix}
    \wt P \\ \wt Q
\end{bmatrix} , \mathcal{N} \left( \begin{bmatrix} \mathds{1}_n \\ 0 \end{bmatrix}, \sigma^2 I_{2n} \right) \right)
$$
and also, by Thm.~\ref{thm:gaussian_data}, we have that 
$$
\mathrm{TV}(\wh P,N_1) + \mathrm{TV}(\wh Q,N_0) \rightarrow 0
$$
as $d \rightarrow \infty$. Moreover, by Thm.~\ref{thm:bin_convergence} and the assumption that $n = \omega(1)$ as a function of $d$, we have that  $H_{\ee^{\veps}}( N_0 , N_1 ) \rightarrow 0$ as $n \rightarrow \infty$, for an appropriate choice of the bid width $h$.
Thus, the claim follows from the inequality~\eqref{eq:ineq_z}.
\end{proof}

\end{cor}

\section{Proof of Lemma~\ref{lem:invertibility}} \label{Asec_lem:invertibility}

\begin{lem}
The hockey-stick divergence $H_\alpha(P_\sigma||Q_\sigma)$ as a function of $\sigma$ is invertible for all $\sigma \in (0,\infty)$.
\begin{proof}
From Eq.~\eqref{eq:delta_gaussian} we know that
\begin{equation} \label{eq:F_dot}
\begin{aligned}
F_\alpha'(\sigma) & = \left( - \log \alpha - \frac{1}{2\sigma^2} \right) \cdot f\left( - \sigma \log \alpha + \frac{1}{2\sigma} \right)  - \alpha \cdot \left( - \log \alpha + \frac{1}{2\sigma^2} \right) \cdot f \left( - \sigma \log \alpha - \frac{1}{2\sigma} \right),
\end{aligned}
\end{equation}
where $f$ denotes the density function of the standard univariate Gaussian distribution.
We see from Eq.~\eqref{eq:F_dot} that for $\alpha=1$, the value of $F_\alpha'(\sigma)$ is strictly negative for all $\sigma>0$. We also know that if a post-processing function reduces the total variation distance, it reduces then all other hockey-stick divergences, since the contraction constant of all hockey-stick divergences is bounded by the contraction constant of the total variation distance. This follows from the fact that for any Markov kernel $\mathrm{K}$, and for any pair of distributions $(P,Q)$ and for any $f$-divergence $D_f( \cdot || \cdot )$, we have that $D_f(P \mathrm{K} ||Q \mathrm{K}) \leq \eta_{\mathrm{TV}}(K) \cdot D_f(P ||Q )$ (see, e.g., Lemma 1 and Thm.\;1 in~\cite{asoodeh2021local}), where $\eta_{\mathrm{TV}}(K)$ denotes the contraction constant of $K$ for the TV distance, i.e., 
$\eta_{\mathrm{TV}}(K) = \sup_{P,Q, \, \mathrm{TV}(P,Q) \neq 0} \frac{\mathrm{TV}(P\mathrm{K},Q\mathrm{K}) }{\mathrm{TV}(P,Q)}$.
Therefore, $F_\alpha'(\sigma)$ is strictly negative for all $\alpha \geq 0$.
\end{proof}
\end{lem}

\section{Proof of Lemma~\ref{lem:max_sens}} \label{Asec_lem:max_sens_proof}

\begin{lem} 
For any $\sigma > 1$, as a function of $\alpha$, $\abs{F_\alpha'(\sigma)}$ has its maximum on the interval $[1,\ee^{\frac{1}{2 \sigma}}]$.
\begin{proof}
The proof goes by looking at the expression $\frac{\dd}{\dd \alpha} F_\alpha'(\sigma)$. Clearly $F_\alpha'(\sigma)$ is negative for all $\sigma>0$ and for all $\alpha \geq 0$.
Thus $\abs{F_\alpha'(\sigma)} = - F_\alpha'(\sigma)$.

Using the expression~\eqref{eq:F_dot}, a lengthy calculation shows that
\begin{equation} \label{eq:alpha_der}
	\begin{aligned}
	 \frac{\dd}{\dd \alpha}  F_\alpha'(\sigma) & =  \frac{1}{\sqrt{2 \pi}}\ee^{-\frac{1}{2}(\frac{1}{2\sigma}  + \log \alpha)^2} \bigg( \bigg(\frac{1}{2\sigma} +\log \alpha \bigg)\bigg(\frac{1}{2\sigma^2}-\log \alpha \bigg)  - \bigg(\frac{1}{2\sigma^2}-\log \alpha \bigg) +1   \bigg)  \\ &+ \frac{1}{\sqrt{2 \pi}}\ee^{-\frac{1}{2}(\frac{1}{2\sigma} - \log \alpha)^2} \bigg( \bigg(- \frac{1}{2\sigma^2} - \log \alpha \bigg) \cdot  \bigg(\frac{1}{2\sigma}-\log \alpha \bigg) - \frac{1}{\alpha} \bigg).
	\end{aligned}
\end{equation}
When $\alpha=1$, i.e., $\log \alpha = 0$, we find from Eq.~\eqref{eq:alpha_der} that
$$
\frac{\dd}{\dd \alpha} F_\alpha'(\sigma)|_{\alpha = 1} = - \frac{1}{2 \sqrt{2  \pi }  \sigma^2}\ee^{ - \frac{1}{8\sigma^2}} < 0.
$$
On the other hand, when $\log \alpha = \frac{1}{2 \sigma}$,  we see from Eq.~\eqref{eq:alpha_der} that
\begin{equation*}
	\begin{aligned}
\frac{\dd}{\dd \alpha}  F_\alpha'(\sigma)|_{\alpha} 
&=  \exp\big(\frac{1}{2 \sigma}\big) = \frac{1}{\sqrt{2 \pi}} \ee^{-\frac{1}{2 \sigma^2}} \big( \frac{1}{\sigma}\big( \frac{1}{2 \sigma^2} - \frac{1}{ \sigma}  \bigg)  - \big( \frac{1}{2 \sigma^2} - \frac{1}{\sigma} \bigg)  \bigg)   + \frac{1}{\sqrt{2 \pi}} \big( \ee^{-\frac{1}{2 \sigma^2}} - \ee^{-\frac{1}{2 \sigma}} \bigg) \\
&= \frac{1}{\sqrt{2 \pi}} \ee^{-\frac{1}{2 \sigma^2}} \big( 1- \frac{1}{\sigma}\bigg) \big(\frac{1}{\sigma} - \frac{1}{2 \sigma^2}  \bigg)  + \frac{1}{\sqrt{2 \pi}} \big( \ee^{-\frac{1}{2 \sigma^2}} - \ee^{-\frac{1}{2 \sigma}} \bigg)
	\end{aligned}
\end{equation*}
which shows that $\frac{\dd}{\dd \alpha} F_\alpha'(\sigma)|_{\alpha = \exp\big(\frac{1}{2 \sigma}\big)} > 0$ when $\sigma > 1$.

Moreover, we can infer from Eq.~\eqref{eq:alpha_der} that $\frac{\dd}{\dd \alpha} F_\alpha'(\sigma)$ is negative when $0 \leq \alpha < 1$ and positive for $\alpha > \ee^{\frac{1}{2 \sigma}}$. Thus $\frac{\dd}{\dd \alpha} \abs{F_\alpha'(\sigma)} = - \frac{\dd}{\dd \alpha} F_\alpha'(\sigma)$ has its maximum 
on the interval $[1,\ee^{\frac{1}{2 \sigma}}]$.

\end{proof}
\end{lem}

\newpage

\section{Algorithm for White-Box Auditing} \label{sec:A_audit}

The following algorithm is considered in the white-box auditing experiments of~\cite{nasr2023tight} and also in the experiments in our Section~\ref{sec:experiments_whitebox}.

\begin{algorithm}[h!]
\caption{White-Box Auditing Using Random Canaries}
\begin{algorithmic}
\STATE{\textbf{Input:} Training dataset $D$, sampling rate $q$, learning rate $\eta$, noise scale $\sigma$, gradient clipping constant $C$, loss function $\ell$, canary gradient $g'$, canary sampling rate $q_c$, function $\mathrm{clip}( \cdot $ that clips vectors to max 2-norm $\wt C$, number of observations $T$, number of training iterations $\tau$.}
\STATE{Observations: $O \rightarrow [], O' \rightarrow []$.}
\STATE{Observations: $O \rightarrow [], O' \rightarrow []$.}
\STATE{Set: $D' = D \bigcup \{ (x',y') \}$.}
\STATE{Initialize: $\theta = \theta_0$.}

\FOR{ $t \in [T]:$ }

        \STATE{$B_t \rightarrow$ Poisson subsample instances from $D$, each with probability $q$.}
        \STATE{$B_t' \rightarrow$ Poisson subsample instances from $D$, each with probability $q$.}
        \STATE{$\nabla[t] \rightarrow \sum_{(x,y) \in B_t}  \mathrm{clip}\big( \nabla_\theta \ell\big( \theta, (x,y) \big) \big)$.}
        \STATE{$\nabla[t] \rightarrow \nabla[t] + \mathcal{N}(0,C^2 \sigma^2)$.}
        \STATE{$\nabla[t]' \rightarrow \sum_{(x,y) \in B_t'}  \mathrm{clip}\big( \nabla_\theta \ell\big( \theta, (x,y) \big) \big)$.}
        \STATE{$\nabla[t]' \rightarrow \nabla[t]' + \mathcal{N}(0,C^2 \sigma^2)$.}
        \STATE{With probability $q_c$: $\nabla[t]' \rightarrow \nabla[t]' + g'$ (add canary with probability $q_c$).}
        \STATE{$O[t] \rightarrow \langle \nabla[t], g' \rangle.$}
        \STATE{$O[t]' \rightarrow \langle \nabla[t]', g' \rangle.$}
        \STATE{$\theta \rightarrow \theta - \eta \nabla[t]$.}
\ENDFOR
\RETURN{$O,O'$.}
\end{algorithmic}
\label{alg:whitebox_auditing}
\end{algorithm}

\end{document}